\documentclass{article}

\PassOptionsToPackage{numbers, sort&compress}{natbib}

\usepackage[preprint] {nips_2018}

\usepackage[utf8]{inputenc} 
\usepackage[T1]{fontenc}    
\usepackage{hyperref}       
\usepackage{url}            
\usepackage{booktabs}       
\usepackage{amsfonts}       
\usepackage{nicefrac}       
\usepackage{microtype}      
\usepackage{graphicx}
\usepackage{subfigure}
\usepackage{algorithm, algorithmicx,algpseudocode}

\usepackage{bbold}
\usepackage{mathtools} 
\usepackage{amsmath,amssymb}

\usepackage{tabularx,ragged2e,booktabs}
\newcolumntype{C}[1]{>{\Centering}m{#1}}

\usepackage{enumitem}
\renewcommand\citet[1]{\cite{#1}}

\usepackage{caption}

\newcommand{\newterm}[1]{{\bf #1}}

\def\secref#1{section~\ref{#1}}

\def\1{\bm{1}}

\def\eps{{\epsilon}}

\DeclareMathAlphabet{\mathsfit}{\encodingdefault}{\sfdefault}{m}{sl}
\SetMathAlphabet{\mathsfit}{bold}{\encodingdefault}{\sfdefault}{bx}{n}

\newcommand{\E}{\mathbb{E}}

\title{Adversarial Examples that Fool both Computer Vision and Time-Limited Humans}

\author{
  Gamaleldin F. Elsayed\thanks{Work done as a member of the Google AI Residency program \href{https://g.co/airesidency}{(g.co/airesidency)}.} \\
  Google Brain\\
  \texttt{gamaleldin.elsayed@gmail.com} \\
 \And
 Shreya Shankar \\
 Stanford University\\
  \And
 Brian Cheung \\
 UC Berkeley\\
  \And
 Nicolas Papernot \\
 Pennsylvania State University\\
   \And
Alex Kurakin \\
Google Brain\\
   \And
Ian Goodfellow \\
Google Brain\\
   \And
Jascha Sohl-Dickstein \\
Google Brain\\
 \texttt{ jaschasd@google.com}\\
}

\begin{document}

\maketitle

\begin{abstract}
Machine learning models are vulnerable to \newterm{adversarial examples}: small changes to images can cause computer vision models to make mistakes such as identifying a school bus as an ostrich. However, it is still an open question whether humans are prone to similar mistakes.
Here, we 
address this question by leveraging recent techniques that transfer adversarial examples from computer vision models with known parameters and architecture to other models with unknown parameters and architecture, and by matching the initial processing of the human visual system. 
We find that adversarial examples that strongly transfer across computer vision models influence the classifications made by time-limited human observers.

\end{abstract}

\section{Introduction}

Machine learning models are easily fooled by adversarial examples: inputs optimized by an adversary to produce an incorrect model classification \citep{szegedy2013intriguing, Biggio13}. In computer vision, an adversarial example is usually an image formed by making small perturbations to an example image. 
Many 
algorithms for constructing adversarial examples \citep{szegedy2013intriguing, goodfellow2014explaining,papernot2015limitations,kurakin17physical,madry2017towards} rely on access to both the architecture and the parameters of the model to perform gradient-based optimization on the input. Without similar access to the brain, these methods do not seem
applicable to constructing adversarial examples for humans.

One interesting phenomenon is that adversarial examples often transfer from one model to another,
making it possible to attack models that an attacker has no access to \cite{szegedy2013intriguing,liu2016delving}.
This naturally raises the question of whether humans
are
susceptible to these adversarial examples. Clearly, humans are prone to many cognitive biases and optical illusions \cite{hillis2002combining}, but these generally do not resemble small perturbations of natural images, nor are they currently generated by optimization of a ML loss function. 
Thus the current understanding 
is that this class of transferable adversarial examples has no effect on human visual perception, yet no thorough empirical investigation has yet been performed.

A rigorous investigation of the above question creates an opportunity both for machine learning to gain knowledge from neuroscience, and for neuroscience to gain knowledge from machine learning.
Neuroscience has often provided existence proofs for machine learning---before we had working object recognition algorithms, we hypothesized it should be possible to build them
because the human brain can recognize objects.
See Hassabis et al.~\cite{hassabis2017neuroscience} for a review of the
influence of neuroscience on artificial intelligence.
If we knew conclusively that the human brain could resist a certain class of adversarial examples, this would provide an existence proof for a similar mechanism in machine learning security. If we knew conclusively that the brain can be fooled by adversarial examples, then machine learning security research should perhaps shift its focus from designing models that are robust to adversarial examples \citep{szegedy2013intriguing, goodfellow2014explaining, papernot2016distillation, xu2017feature, ensemble_training, madry2017towards, kolter2017provable, buckman2018thermometer} to designing systems that are secure despite including non-robust machine learning components. Likewise, if adversarial examples developed for computer vision affect the brain, this phenomenon discovered in the context of machine learning could lead to a better understanding of brain function.

In this work, we construct adversarial examples that transfer from computer vision
models to the human visual system.
In order to successfully construct these examples and observe their effect,
we leverage three key ideas from machine learning, neuroscience, and psychophysics.
First, we use the recent \newterm{black box} adversarial example construction techniques
that create adversarial examples for a target model without access to the model's architecture or parameters. Second, we adapt machine learning models to mimic the initial visual processing of humans,
making it more likely that adversarial examples will transfer from the model to a human observer.
Third, we evaluate classification decisions of human observers in a time-limited setting,
so that even subtle effects on human perception are detectable.
By making image presentation sufficiently brief, 
humans are unable to achieve perfect accuracy even on clean images, and small changes in 
performance lead to more measurable changes in accuracy. 
Additionally, a brief image presentation 
limits the time in which the brain can utilize recurrent and top-down processing pathways \cite{potter2014detecting}, 
and is believed to make the processing in the brain more closely resemble that in a 
feedforward artificial neural network.

We find that adversarial examples that transfer across computer vision models {\em do} successfully influence the perception of human observers, thus uncovering a new class of illusions that are shared between computer vision models and the human brain.

\section{Background and Related Work}

\subsection{Adversarial Examples}

Goodfellow et al.~\cite{goodfellow2017} define adversarial examples as ``inputs to machine learning models
that an attacker has intentionally designed to cause the model to make a mistake.''
In the context of visual object recognition, adversarial examples are images
usually formed by applying a small perturbation to a naturally occurring image
 in a way that breaks the predictions made by a machine learning classifier.
 See Figure \ref{fig: illustration}a for a canonical example where adding a
 small perturbation to an image of a panda causes it to be misclassified as a gibbon. This perturbation is small enough to be imperceptible (i.e., it cannot be saved in a standard png file that uses 8 bits because the perturbation is smaller than $1/255$ of the pixel dynamic range). This perturbation relies on carefully chosen structure based on the parameters of the neural network---but when magnified to be perceptible, human observers cannot recognize any meaningful structure. Note that adversarial examples also exist in other domains like malware detection \citep{grosse17}, but we focus here on image classification tasks.

Two aspects of the definition of adversarial examples are particularly important
for this work:
\begin{enumerate}
    \item Adversarial examples are designed to cause a {\em mistake}.
    They are not (as is commonly misunderstood) defined to {\em }differ from human judgment.
    If adversarial examples were defined by deviation from human output, it would
    by definition be impossible to make adversarial examples for humans.
    On some tasks, like predicting whether input numbers are prime, there is a clear
    objectively correct answer, and we would like the model to get the correct
    answer, not the answer provided by humans (time-limited humans are probably not
    very good at guessing whether numbers are prime).
    It is challenging to define what constitutes a mistake for visual
    object recognition.
    After adding a perturbation to an image it likely no longer corresponds to a photograph of 
    a real physical scene.
    Furthermore, 
    it is philosophically difficult to 
    define the real object class for an image that is not a picture of a real object. 
    In this work, we assume that an adversarial image is misclassified if the output
    label differs from the human-provided label of the clean image that was used as
    the starting
    point for the adversarial image. We make small adversarial perturbations
    and we assume that these small perturbations are insufficient to change the true
    class.
    \item Adversarial examples are not (as is commonly misunderstood) defined to be imperceptible. If this were the case, it would be impossible by definition to make adversarial examples for
    humans, because changing the human's classification would constitute a change in what
    the human perceives (e.g., see Figure \ref{fig: illustration}b,c).
\end{enumerate}

\subsubsection{Clues that Transfer to Humans is Possible}

Some observations give clues that 
transfer to humans may be possible. Adversarial examples are known to transfer across machine learning models, which suggest that these adversarial perturbations may carry information about target adversarial classes. Adversarial examples that fool one model often fool another model with a different
architecture~\cite{szegedy2013intriguing}, another model that was trained on a different training set~\cite{szegedy2013intriguing}, or even trained with a different algorithm~\cite{papernot2016transferability}
 (e.g.,
adversarial examples designed to fool a convolution neural network
may also fool a decision tree). 
The transfer effect makes it possible to perform
black box attacks, where adversarial examples fool models that an attacker does not have access to
\citep{szegedy2013intriguing,papernot2017practical}. Kurakin et al.~\cite{kurakin17physical} found that adversarial examples transfer from the digital to the physical world, despite many transformations such as lighting and camera effects that modify their appearance when they are photographed in the physical world. Liu et al.~\cite{liu2016delving} showed that the transferability of an adversarial example can be greatly improved by optimizing it to fool {\em many} machine learning models rather than one model: an adversarial example that fools five models used in the optimization process is more likely to fool an arbitrary sixth model.

\begin{figure}[t]
\vskip 0.05in
\begin{center}
\centerline{\includegraphics[width=\columnwidth]{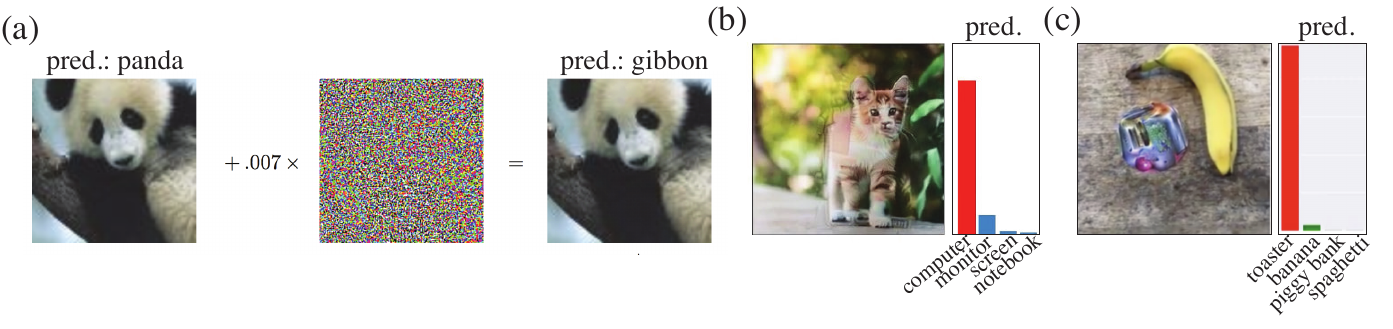}}
\captionof{figure}{
\textbf{Adversarial examples optimized on more models / viewpoints sometimes appear
more meaningful to humans.}
This observation is a clue that machine-to-human transfer may be possible.
(a) A canonical example of an adversarial image reproduced from \citet{goodfellow2014explaining}.
This adversarial attack 
has moderate but limited ability to fool the model after geometric transformations
or to fool models other than the model used to generate the image.
(b) An adversarial attack causing a cat image to be labeled as a computer while being robust to geometric transformations, adopted from \citet{athalye2017blog}. Unlike the attack in a, the image contains features that seem semantically computer-like to humans. (c) An adversarial patch that causes images to be labeled as a toaster, optimized to cause misclassification from multiple viewpoints, reproduced from \citet{brown2017adversarial}. Similar to b, the patch contains features that appear toaster-like to a human.
}
\label{fig: illustration}
\end{center}
\vskip -0.2in
\end{figure}

Moreover, recent studies on stronger adversarial examples that 
transfer across 
multiple settings 
have sometimes produced 
adversarial examples that
appear more meaningful to human observers. For instance, a cat adversarially
perturbed to resemble a computer~\cite{athalye2017synthesizing} while
transfering across geometric transformations develops features that appear
computer-like (Figure \ref{fig: illustration}b), and the `adversarial toaster'
from Brown et al.~\cite{brown2017adversarial} possesses features that seem
toaster-like (Figure \ref{fig: illustration}c).
This development of human-meaningful features is consistent with the adversarial 
example carrying true feature information and thus coming closer to fooling humans, if we acounted for the notable differences between humans visual processing and computer vision models (see~\secref{sec: Notable Differences})

\subsection{Biological and Artificial Vision}

\subsubsection{Similarities}

Recent research has found
similarities in
representation and behavior between deep convolutional neural networks (CNNs) and the primate visual system \citep{cadieu2014deep}. This further motivates the possibility that adversarial examples may transfer from computer vision models to humans. Activity in deeper CNN layers has been observed to be predictive of activity recorded in the visual pathway of primates~\cite{cadieu2014deep,Yamins2016UsingGD}. 
Reisenhuber and Poggio~\cite{riesenhuber1999hierarchical} developed a model of object recognition in cortex that closely resembles many aspects of modern CNNs.  Kummerer et al.~\cite{kummerer2014deep, kummerer2017deepgaze} showed that CNNs are predictive of human gaze fixation. Style transfer \citep{gatys2015neural} demonstrated that intermediate layers of a CNN capture notions of artistic style which are meaningful to humans. Freeman et al.~\cite{freeman2011metamers} used representations in a CNN-like model to develop psychophysical metamers, which are indistinguishable to humans when viewed briefly and with carefully controlled fixation. 
Psychophysics experiments have compared the pattern of errors made by humans, to that made by neural network classifiers~\cite{geirhos2017comparing, Rajalingham240614}.

\subsubsection{Notable Differences}
\label{sec: Notable Differences}
Differences between machine and human vision occur early in the visual system. Images are typically presented to CNNs as a static rectangular pixel grid with constant spatial resolution. The primate eye on the other hand has an eccentricity dependent spatial resolution. Resolution is high in the fovea, or central $\sim$ $5^{\circ}$ of the visual field, but falls off linearly with increasing eccentricity \citep{retinal}. A perturbation which requires high acuity in the periphery of an image, as might occur as part of an adversarial example, would be 
undetectable by the eye, 
and thus would have no impact on human perception. 
Further differences include the sensitivity of the eye to 
temporal as well as spatial features, 
as well as non-uniform color sensitivity 
\citep{land2012animal}. Modeling the early visual system continues to be an area of active study \citep{olshausen201320,mcintosh2016deep}. As we describe in \secref{ensemble_of_models}, we mitigate some of these differences by using a biologically-inspired
image 
input layer.

Beyond early visual processing, there are more major computational differences between CNNs and the human brain. 
All the CNNs we consider are fully feedforward architectures, while the visual cortex has many times more feedback than feedforward connections, as well as extensive recurrent dynamics \citep{olshausen201320}. 
Possibly due to these differences in architecture, humans have been found experimentally to make classification mistakes that are qualitatively different than those made by deep networks \cite{eckstein2017humans}. 
Additionally, the brain does not treat a scene as a single static image, but
actively explores it
with saccades
\citep{ibbotson2011visual}. As is common in psychophysics experiments \citep{kovacs1995cortical}, we
mitigate these differences in processing by limiting both the way in which the image is presented, and the time which the subject has to process it, as described in \secref{sec: human experiment}.

\section{Methods}\label{sec methods}
Section \ref{sec: image generation pipline} details
our 
machine learning vision pipeline. Section \ref{sec: human experiment}
describes our psychophysics experiment to evaluate the impact of adversarial images on human subjects.

\subsection{The Machine Learning Vision Pipeline}
\label{sec: image generation pipline}
\subsubsection{Dataset}

In our experiment, we used images from ImageNet \cite{deng2009imagenet}.
ImageNet contains 1,000 highly specific
classes that typical people may not be able
to identify, such as ``Chesapeake Bay retriever''.
Thus, we combined some of these fine classes to
form six coarse classes we were confident would
be familiar to our experiment subjects ($\{$dog, cat, broccoli, cabbage, spider, snake$\}$). We then grouped these six classes into the following groups: (i) \newterm{Pets} group (dog and cat images); (ii) \newterm{Hazard} group (spider and snake images); (iii) \newterm{Vegetables} group (broccoli and cabbage images).

\subsubsection{Ensemble of Models}
\label{ensemble_of_models}

We constructed an ensemble of $k$ CNN models trained on ImageNet ($k=10$). Each model is an instance of one of these
architectures: Inception V3, Inception V4, Inception ResNet V2, ResNet V2 50, ResNet V2 101, and ResNet V2 152 \citep{inceptionv3, inceptionresnet, resnet}. To better match the initial processing of human visual system, we prepend each model with a retinal layer, which pre-processes the input to incorporate some of the transformations performed by the human eye. In that layer, we perform an eccentricity dependent blurring of the image to approximate the input which is received by the visual cortex of human subjects through their retinal lattice. 
The details of this retinal layer are described in Appendix \ref{sec app retinal}. 
We use eccentricity-dependent spatial resolution measurements (based on the macaque visual system)~\cite{retinal}, along with the known geometry of the viewer and the screen, to determine the
degree of spatial blurring at each image location. 
This limits the CNN to information which is also available to the human visual system.
The layer is fully differentiable, allowing gradients to backpropagate through the network when running adversarial attacks.
Further details of the models and their classification performance
are provided in Appendix \ref{sec: models}.

\subsubsection{Generating Adversarial Images}
\label{sec generate}

For a given image group, we wish to generate targeted adversarial examples that strongly transfer across models. This means that for a class pair $(A, B)$ (e.g., $A$: cats and $B$: dogs), we generate adversarial perturbations such that models will classify perturbed images from $A$ as $B$; similarly, we perturbed images from $B$ to be classified as $A$. A different perturbation is constructed for each image; however, the $\ell_{\infty}$ norm of all perturbations are constrained to be equal to a fixed $\epsilon$. 

Formally: given a classifier $k$, which assigns probability $P (y \mid X)$ to each coarse class $y$ given an input image $X$, a specified target class $y_\mathrm{target}$ and a maximum perturbation $\epsilon$, we want to find the image $X_{adv}$ that minimizes $-\log (P (y_\mathrm{target} \mid X_{adv}))$ with the constraint that $||X_{adv} - X||_{\infty} = \epsilon$. See Appendix \ref{sec: coarse prob} for details on computing the coarse class probabilities $P (y \mid X)$. With the classifier's parameters, we can perform iterated gradient descent on $X$ in order to generate our $X_{adv}$ (see Appendix \ref{sec: adv gen}). This iterative approach is commonly employed to generate adversarial images~\cite{kurakin17physical}.

\begin{figure}[t]
\vskip 0.05in
\begin{center}
\centerline{\includegraphics[width=\columnwidth]{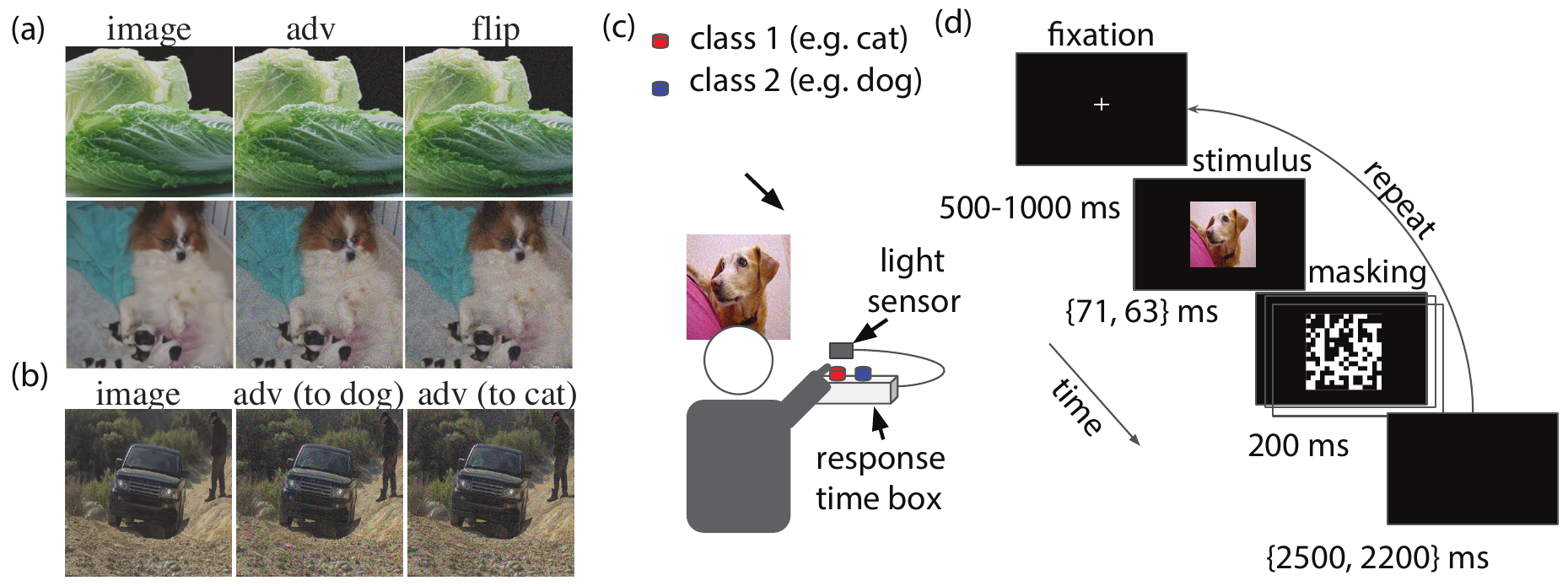}}
\caption{\textbf{Experiment setup and task.} 
(a)  examples images from the conditions (\texttt{image}, \texttt{adv}, and \texttt{flip}). Top: \texttt{adv} targeting broccoli class. bottom: \texttt{adv} targeting cat class. See definition of conditions at Section \ref{sec: experiment conditions}. (b) example images from the  \texttt{false} experiment condition. (c) Experiment setup and recording apparatus. (d) Task structure and timings. The subject is asked to repeatedly identify which of two classes (e.g. dog vs. cat) a briefly presented image belongs to. The image is either adversarial, or belongs to one of several control conditions. See Section \ref{sec: human experiment} for details.}
\label{fig:experiment}
\end{center}
\vskip -0.2in
\end{figure}
\subsection{Human Psychophysics Experiment}
\label{sec: human experiment}
38 subjects with normal or corrected vision participated in the experiment. Subjects gave informed consent to participate, and were awarded a reasonable compensation for their time and effort\footnote{The study was granted an Institutional Review Board (IRB) exemption by an external, independent, ethics board (Quorum review ID 33016).}.

\subsubsection{Experimental Setup}
\label{sec exp setup}

Subjects sat on a fixed chair $61$cm away from a high refresh-rate computer screen (ViewSonic XG2530) in a room with dimmed light
(Figure \ref{fig:experiment}c).
Subjects were asked to classify images that appeared on the screen to one of two classes (two alternative forced choice) by pressing buttons on a response time box (LOBES v5/6:USTC) using two fingers on their right hand. The assignment of classes to buttons was randomized for each experiment session. Each trial started with a fixation cross displayed in the middle of the screen for $500-1000$ ms, instructing subjects to direct their gaze to the fixation cross (Figure \ref{fig:experiment}d). After the fixation period, an image of size $15.24\text{ cm} \times 15.24\text{ cm}$ ($14.2^{\circ}$ visual angle) was presented briefly at the center of the screen for a period of $63$ ms ($71$ ms for some sessions). The image was followed by a sequence of ten high contrast binary random masks, each displayed for $20$ ms (see example in Figure \ref{fig:experiment}d). Subjects were asked to classify the object in the image (e.g., cat vs. dog) by pressing one of two buttons starting at the image presentation time and lasting until $2200$ ms (or $2500$ ms for some sessions) after the mask was turned off. The waiting period to start the next trial was of the same duration whether subjects responded quickly or slowly. Realized exposure durations were $\pm 4$ms from the times reported above, as measured by a photodiode and oscilloscope in a separate test experiment. Each subject's response time was recorded by the response time box relative to the image presentation time (monitored by a photodiode). In the case
where a subject pressed more than one button in a trial, only the class corresponding to their first choice was considered. Each subject completed between 140 and 950 trials.

\subsubsection{Experiment Conditions}
\label{sec: experiment conditions}

Each experimental session included only one of the image groups (Pets, Vegetables or Hazard). For each group, images were presented in one of four conditions as follows:

\begin{itemize}
\item \texttt{image}: images from the ImageNet training set (rescaled to the [$40, 255-40$] range to avoid clipping when adversarial perturbations are added; see Figure \ref{fig:experiment}a left) ).
\item \texttt{adv}: we added adversarial perturbation $\delta_{adv}$ to \texttt{image}, crafted to cause machine learning models to  misclassify  \texttt{adv} as the opposite class in the group (e.g., if \texttt{image} was originally a cat, we perturbed the image to be classified as a dog). We used a perturbation size large enough to be noticeable by humans on the computer screen but small with respect to the image intensity scale ($\epsilon=32$; see Figure \ref{fig:experiment}a middle).
In other words, we chose $\epsilon$ to be large (to improve the chances of 
adversarial examples transfer to time-limited human) but kept it small enough that the perturbations
are class-preserving (as judged by a no-limit human).
\item \texttt{flip}: similar to \texttt{adv}, but 
the adversarial perturbation ($\delta_{adv}$) is flipped vertically before being added to \texttt{image}. This is a control condition, 
chosen to have nearly identical perturbation statistics to the \texttt{adv} condition (see Figure \ref{fig:experiment}a right).
We include this condition because 
if adversarial perturbations {\em reduce the accuracy}
of human observers, this could just be because the
perturbations degrade the image quality.
\item \texttt{false}:
in this condition, subjects are forced to make a mistake.
To show that adversarial perturbations
{\em actually control the chosen class},
we include this condition where neither of the two 
options available to the subject is correct,
so their accuracy is always zero. We test
whether adversarial perturbations can influence which
of the two wrong choices they make.
We show a random image from an ImageNet class other than the two classes in the group, and adversarially perturb it
toward one of the two classes in the group.
The subject must then choose from these two classes.
For example, we might show an airplane 
adversarially 
perturbed toward the dog class, while a subject is 
in a session classifying images as cats or dogs.
We used a slightly larger perturbation in this condition ($\epsilon=40$; see Figure \ref{fig:experiment}b).
\end{itemize}

The conditions (\texttt{image}, \texttt{adv}, \texttt{flip}) are ensured to have balanced number of trials within a session by either uniformly sampling the condition type in some of the sessions or randomly shuffling a sequence with identical trial counts for each condition in other sessions. The number of trials for each class in the group was also constrained to be equal. Similarly for the \texttt{false} condition the number of trials adversarially perturbed towards class 1 and class 2 were balanced for each session. 
To reduce subjects using strategies based on overall color or brightness distinctions between classes, we pre-filtered the dataset to remove images that showed an obvious effect of this nature. Notably, in the pets group we excluded images that included large green lawns or fields, since in almost all cases these were photographs of dogs. See Appendix \ref{app image list} for images used in the experiment for each coarse class. 
For example images for each condition, see Figures 
 \ref{fig advx false} through 
\ref{fig advx veg}.

\section{Results}
\label{sec: results}

\subsection{Adversarial Examples Transfer to Computer Vision Models}
\label{model_eval}

We first assess the transfer of our constructed images to two test models that were not included in the ensemble used to generate adversarial examples. These test models are 
an adversarially trained Inception V3 model \cite{kurakin2016mlatscale} and a ResNet V2 50 model. 
Both models perform well ($> 75\%$ accuracy) on clean images. 
Attacks in the \texttt{adv} and \texttt{false} conditions succeeded against the test models between $57\%$ and $89\%$ of the time, depending on image class and experimental condition. The \texttt{flip} condition changed the test model predictions on fewer than $1.5\%$ of images in all conditions, validating its use as a control. 
See Tables 
\ref{table: train accuracy} - \ref{table: test attack success}
for accuracy and attack success measurements on both train and test models for all experimental conditions.

\begin{figure}[t!]
\vskip 0.2in
\begin{center}
\centerline{\includegraphics[width=\columnwidth]{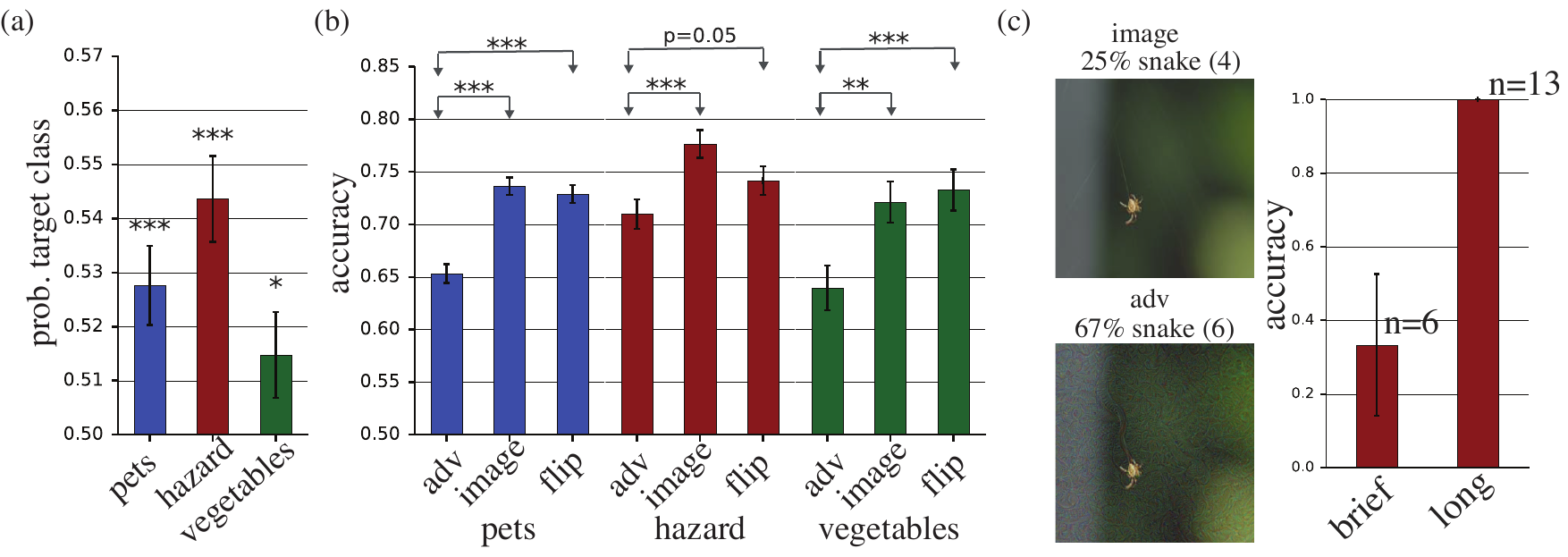}}
\caption{\textbf{Adversarial images transfer to humans.}
(a) By adding adversarial perturbations to an image, we are able to bias which of two incorrect choices subjects make. Plot shows probability of choosing the adversarially targeted class when the true image class is not one of the choices that subjects can report (\texttt{false} condition), estimated by averaging the responses of all subjects (two-tailed t-test relative to chance level $0.5$). (b) Adversarial images cause more mistakes than either clean images or images with the adversarial perturbation flipped vertically before being applied. Plot shows probability of choosing the true image class, when this class is one of the choices that subjects can report, averaged across all subjects. Accuracy is significantly less than 1 even for clean images due to the brief image presentation time. (error bars $\pm$ SE; *: $p<0.05$; **: $p<0.01$; ***: $p<0.001$) (c) A spider image that time-limited humans frequently perceived as a snake 
(top parentheses: number of subjects tested on this image). right: accuracy on this adversarial image when presented briefly compared to when presented for long time (long presentation is based on a post-experiment email survey of 13 participants). 
}
\label{fig: results1}
\end{center}
\vskip -0.2in
\end{figure}

\begin{figure}[t!]
\vskip 0.2in
\begin{center}
\centerline{\includegraphics[width=0.9\linewidth]{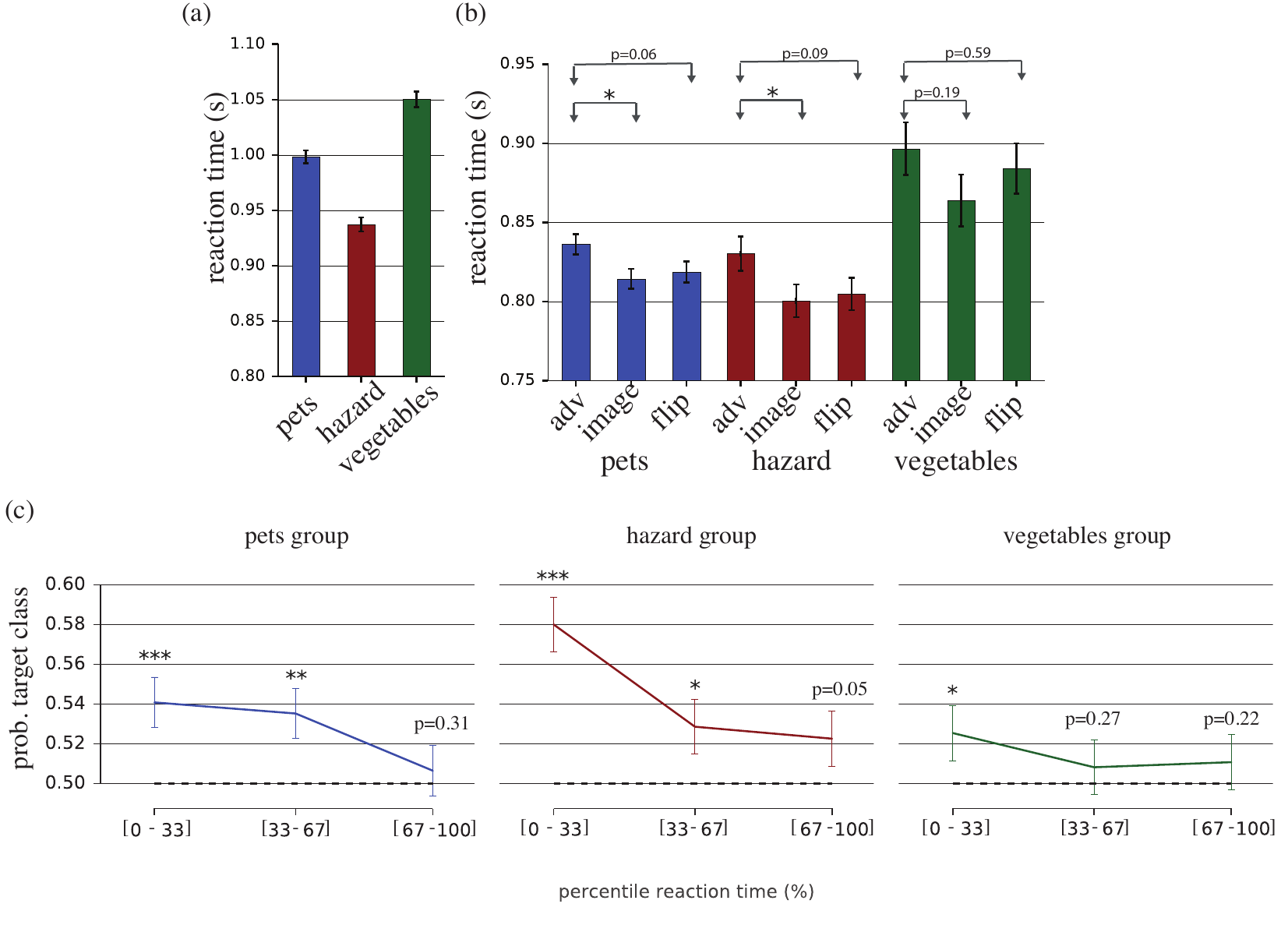}}
\captionof{figure}{\textbf{Adversarial images effect human response time.} 
(a) Average response time to \texttt{false} images. (b) Average response time for \texttt{adv}, \texttt{image}, and \texttt{flip} conditions (error bars $\pm$ SE; * reflects $p<0.05$; two sample two-tailed t-test). 
In all three stimulus groups, there was a trend towards slower response times in the \texttt{adv} condition than in either control group. 
(c) Probability of choosing the adversarially targeted class in the \texttt{false} condition, estimated by averaging the responses of all subjects (two-tailed t-test relative to chance level $0.5$; error bars $\pm$ SE; *: $p<0.05$; **: $p<0.01$; ***: $p<0.001$). The probability of choosing the targeted label is computed by binning trials within percentile reaction time ranges (0-33 percentile, 33-67 percentile, and 67-100 percentile). The bias relative to chance level of 0.5 is significant when people reported their decision quickly (when they may have been more confident), but not significant when they reported their decision more slowly. 
As discussed in Section \ref{transfer to humans}, differing effect directions in (b) and (c) may be explained by adversarial perturbations decreasing decision confidence in the \texttt{adv} condition, and increasing decision confidence in the \texttt{false} condition.
}
\label{fig:results1_rt}
\end{center}
\vskip -0.2in
\end{figure}


\subsection{Adversarial Examples Transfer to Humans}

We now show that adversarial examples transfer to time-limited humans.
One could imagine that adversarial examples merely degrade image quality
or discard information, thus increasing error rate.
To rule out this possibility, we begin by showing that for a fixed
error rate (in a setting where the human is forced to make a mistake),
adversarial perturbations influence the human choice
among two incorrect classes.
Then, we demonstrate that adversarial examples increase
the error rate.

\subsubsection{Influencing the Choice between two Incorrect Classes}
\label{human_eval}

As described in Section \ref{sec: experiment conditions},
we used the \texttt{false} condition to test whether
adversarial perturbations can influence which of two
incorrect classes a subject chooses (see example images in Figure \ref{fig advx false}).

We measured our effectiveness at changing the perception of subjects using the rate at which subjects reported the adversarially targeted class.
If the adversarial perturbation were completely ineffective
we would expect the choice of targeted class to be uncorrelated
with the subject's reported class.
The average rate at which the subject chooses the
target class metric would be $0.5$ as each \texttt{false} image can be perturbed to class 1 or class 2 in the group with equal probability. Figure \ref{fig: results1}a shows the probability of choosing the target class averaged across all subjects for all the three experiment groups. In all cases, the probability was significantly above the chance level of $0.5$. This demonstrates that the adversarial perturbations generated using CNNs biased human perception towards the targeted class. This effect was stronger for the the hazard, then pets, then vegetables group. This difference in probability among the class groups was significant ($p<0.05$; Pearson Chi-2 GLM test).

We also observed a significant difference in the mean response time between the class groups ($p<0.001$; one-way ANOVA test; see Figure \ref{fig:results1_rt}a). Interestingly, the response time pattern across image groups (Figure \ref{fig:results1_rt}a)) was inversely correlated to the perceptual bias pattern (Figure \ref{fig: results1}a)) (Pearson correlation $= -1$, $p<0.01$; two-tailed Pearson correlation test). In other words, subjects made quicker decisions for the hazard group, then pets group, and then vegetables group. This is consistent with subjects being more confident in their decision when the adversarial perturbation was more successful in biasing subjects perception. This inverse correlation between attack success and response time was observed within group, as well as between groups (Figure \ref{fig:results1_rt}).

\subsubsection{Adversarial Examples Increase Human Error Rate}
\label{transfer to humans}
We demonstrated that we are able to bias human perception to a target class when the true class of the image is not one of the options that subjects can choose.
Now we show that adversarial perturbations can be used
to cause the subject to choose an incorrect class
even though the correct class is an available response.
As described in Section \ref{sec: experiment conditions}, we presented \texttt{image}, \texttt{flip}, and \texttt{adv}. 

Most subjects
had lower accuracy in \texttt{adv} than \texttt{image} (Table \ref{table: people count}). This is also reflected on the average  significantly lower accuracy across all subjects  for the \texttt{adv} than \texttt{image} (Figure \ref{fig: results1}b). 

The above result may simply imply that the signal to noise ratio in the adversarial images is lower than that of clean images. 
While this issue is partially addressed with the
\texttt{false} experiment results in Section \ref{human_eval},
we additionally tested accuracy on \texttt{flip} images.
This control case uses perturbations with identical statistics to \texttt{adv} up to a flip of the vertical axis. However, this control breaks the pixel-to-pixel correpsondence between the adversarial perturbation and the image. 
The majority of subjects
had lower accuracy in the \texttt{adv} condition
than in the \texttt{flip} condition
(Table \ref{table: people count}). 
When averaging across all trials, this effect was very significant for the pets and vegetables group ($p<0.001$), and less significant for the hazard group ($p=0.05$) (Figure \ref{fig: results1}b). These results suggest that the direction of the adversarial image perturbation, in combination with a specific image, is perceptually relevant to features that the human visual system uses to classify objects. These findings thus give evidence that strong black box adversarial attacks can transfer from CNNs to humans, and show remarkable similarities between failure cases of CNNs and human vision.

In all cases, the average response time
was longer for the \texttt{adv} condition relative to the other conditions (Figure \ref{fig:results1_rt}b), 
though this result was only statistically significant for two comparisons. 
If this trend remains predictive, it would seem to contradict the case when we presented \texttt{false} images (Figure \ref{fig:results1_rt}a). 
One interpretation is that in the \texttt{false} case, the transfer of adversarial features to humans was accompanied by more confidence,  whereas here the transfer was accompanied by less confidence, possibly due to competing adversarial and true class features in the \texttt{adv} condition.

\begin{figure}[t!]
\vskip 0.2in
\begin{center}
\centerline{\includegraphics[width=\columnwidth]{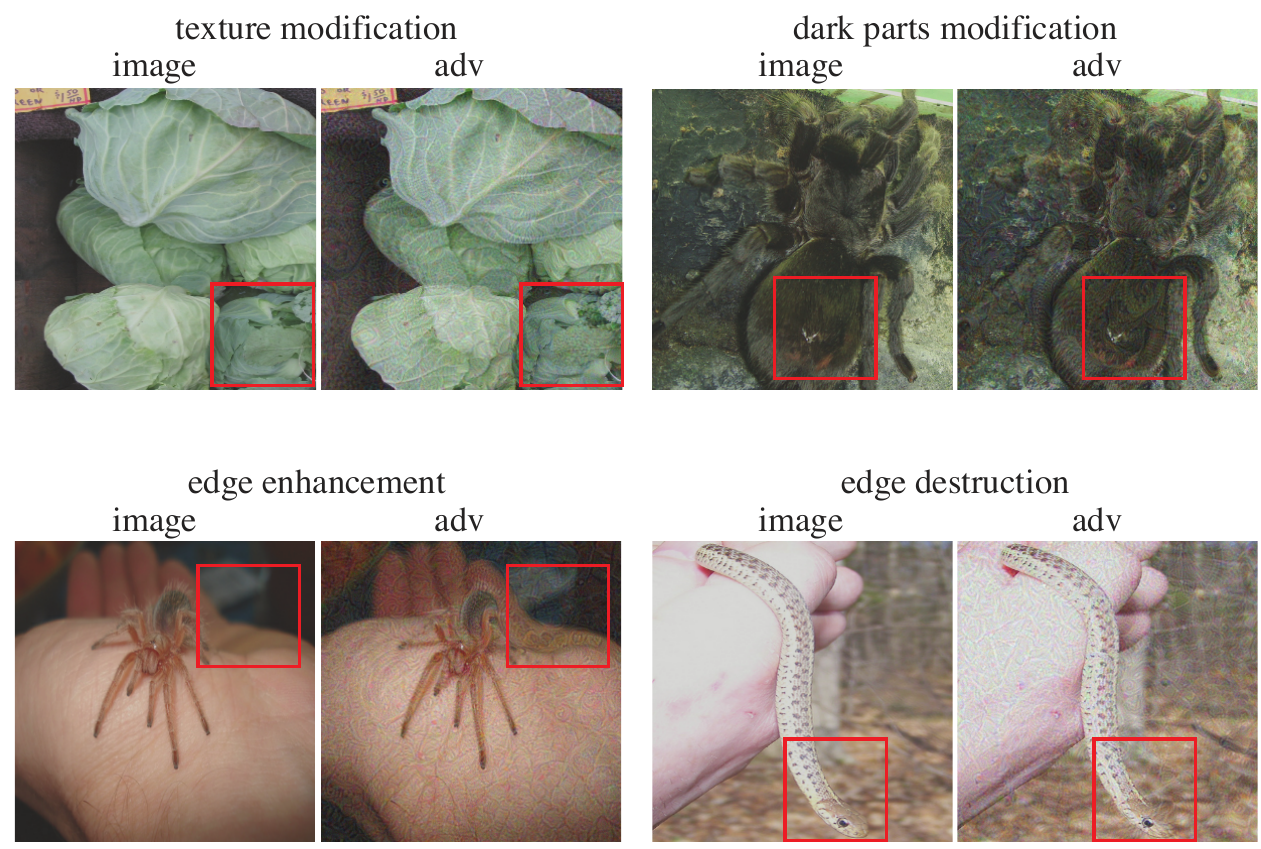}}
\captionof{figure}{\textbf{Examples of the types of manipulations performed by the adversarial attack.} 
See Figures \ref{fig advx pets} through 
\ref{fig advx veg} for additional examples of adversarial images. Also see Figure  \ref{fig advx false} for adversarial examples from the \texttt{false} condition.
}
\label{fig examples}
\end{center}
\vskip -0.2in
\end{figure}

\section{Discussion}
\label{sec: discussion}

Our results invite several questions that we discuss briefly.

\subsection{Have we actually fooled human observers or did we change the true class?}

One might naturally wonder whether we have fooled the human observer or
whether we have replaced the input image with an image that actually
belongs to a different class.
In our work, the perturbations we made were small enough that they generally
do not change the output class for a human who has no time limit
(the reader may verify this by observing Figures \ref{fig:experiment}a,b, \ref{fig: results1}c, and 
 \ref{fig advx false} through \ref{fig advx veg}).

.
We can thus be confident that we did not change the true class of the image,
and that we really did fool the time-limited human.
Future work aimed at fooling humans with no time-limit will need to tackle
the difficult problem of obtaining a better ground truth signal than visual labeling 
by humans.

\subsection{How do the adversarial examples work?}

We did not design
controlled experiments to prove that the adversarial
examples work in any specific way, but we informally observed a few apparent patterns
illustrated in Figure \ref{fig examples}:
disrupting object edges, 
especially by 
mid-frequency modulations perpendicular to the edge; 
enhancing edges both by increasing contrast and creating texture boundaries; 
modifying texture; 
and taking advantage of dark regions in the image, where the perceptual magnitude of small $\epsilon$ perturbations can be larger.

\subsection{What are the implications for machine learning security and society?}

The fact that our transfer-based adversarial examples fool time-limited humans but not
no-limit humans suggests that the lateral and top-down connections used by the no-limit
human are relevant to human robustness to adversarial examples.
This suggests that machine learning security research should explore the significance
of these top-down or lateral connections further.
One possible explanation for our observation is that no-limit humans are fundamentally
more robust to adversarial example and achieve this robustness via top-down or lateral
connections.
If this is the case, it could point the way to the development of more robust machine
learning models.
Another possible explanation is that no-limit humans remain highly vulnerable to adversarial
examples but adversarial examples do not transfer from feed-forward networks to no-limit
humans because of these architectural differences.

Our results suggest that there is a risk that imagery could be manipulated to cause
human observers to have unusual reactions; for example, perhaps a photo of a politician
could be manipulated in a way that causes it to be perceived as unusually untrustworthy
or unusually trustworthy in order to affect the outcome of an election.

\subsection{Future Directions}

In this study, we designed a procedure that according to our hypothesis would transfer adversarial examples to humans. An interesting set of questions relates to how sensitive that transfer is to 
different elements of our experimental design. 
For example:
How does transfer depend on $\epsilon$? Was model ensembling crucial to transfer? Can the retinal preprocessing layer be removed? We suspect that retinal preprocessing and ensembling are both important for transfer to humans, but that $\epsilon$ could be made smaller. See Figure \ref{fig cat dog} for a preliminary exploration of these questions.

\section{Conclusion}

In this work, we showed that adversarial examples based on
perceptible but class-preserving perturbations
that fool
multiple machine learning models also fool time-limited humans.
Our findings demonstrate striking similarities between convolutional
neural networks and the human visual system.
We expect this observation to lead to advances in both neuroscience
and machine learning research.

\section*{Acknowledgements}
We are grateful to 
Ari Morcos, 
Bruno Olshausen, 
David Sussillo, 
Hanlin Tang, 
John Cunningham,
Santani Teng,
and 
Daniel Yamins 
for useful discussions. We also thank 
Dan Abolafia 
Simon Kornblith, 
Katherine Lee, 
Niru Maheswaranathan, 
Catherine Olsson, 
David Sussillo, 
and 
Santani Teng, 
for helpful feedback on the manuscript. We thank Google Brain residents for useful feedback on the work. 
We also thank Deanna Chen, Leslie Philips, Sally Jesmonth, Phing Turner, Melissa Strader, Lily Peng, and Ricardo Prada for assistance with IRB and experiment setup.

\bibliography{biblio}

\begin{thebibliography}{10}

\bibitem{athalye2017blog}
Anish Athalye.
\newblock Robust adversarial examples, 2017.

\bibitem{athalye2017synthesizing}
Anish Athalye and Ilya Sutskever.
\newblock Synthesizing robust adversarial examples.
\newblock {\em arXiv preprint arXiv:1707.07397}, 2017.

\bibitem{Biggio13}
Battista Biggio, Igino Corona, Davide Maiorca, Blaine Nelson, Nedim Srndic,
  Pavel Laskov, Giorgio Giacinto, and Fabio Roli.
\newblock Evasion attacks against machine learning at test time.
\newblock In {\em Machine Learning and Knowledge Discovery in Databases -
  European Conference, {ECML} {PKDD} 2013, Prague, Czech Republic, September
  23-27, 2013, Proceedings, Part {III}}, pages 387--402, 2013.

\bibitem{brown2017adversarial}
Tom~B Brown, Dandelion Man{\'e}, Aurko Roy, Mart{\'\i}n Abadi, and Justin
  Gilmer.
\newblock Adversarial patch.
\newblock {\em arXiv preprint arXiv:1712.09665}, 2017.

\bibitem{buckman2018thermometer}
Jacob Buckman, Aurko Roy, Colin Raffel, and Ian Goodfellow.
\newblock Thermometer encoding: One hot way to resist adversarial examples.
\newblock {\em International Conference on Learning Representations}, 2018.
\newblock accepted as poster.

\bibitem{cadieu2014deep}
Charles~F Cadieu, Ha~Hong, Daniel~LK Yamins, Nicolas Pinto, Diego Ardila,
  Ethan~A Solomon, Najib~J Majaj, and James~J DiCarlo.
\newblock Deep neural networks rival the representation of primate it cortex
  for core visual object recognition.
\newblock {\em PLoS computational biology}, 10(12):e1003963, 2014.

\bibitem{deng2009imagenet}
Jia Deng, Wei Dong, Richard Socher, Li-Jia Li, Kai Li, and Li~Fei-Fei.
\newblock Imagenet: A large-scale hierarchical image database.
\newblock In {\em Computer Vision and Pattern Recognition, 2009. CVPR 2009.
  IEEE Conference on}, pages 248--255. IEEE, 2009.

\bibitem{eckstein2017humans}
Miguel~P Eckstein, Kathryn Koehler, Lauren~E Welbourne, and Emre Akbas.
\newblock Humans, but not deep neural networks, often miss giant targets in
  scenes.
\newblock {\em Current Biology}, 27(18):2827--2832, 2017.

\bibitem{freeman2011metamers}
Jeremy Freeman and Eero~P Simoncelli.
\newblock Metamers of the ventral stream.
\newblock {\em Nature neuroscience}, 14(9):1195, 2011.

\bibitem{gatys2015neural}
Leon~A Gatys, Alexander~S Ecker, and Matthias Bethge.
\newblock A neural algorithm of artistic style.
\newblock {\em arXiv preprint arXiv:1508.06576}, 2015.

\bibitem{geirhos2017comparing}
Robert Geirhos, David~HJ Janssen, Heiko~H Sch{\"u}tt, Jonas Rauber, Matthias
  Bethge, and Felix~A Wichmann.
\newblock Comparing deep neural networks against humans: object recognition
  when the signal gets weaker.
\newblock {\em arXiv preprint arXiv:1706.06969}, 2017.

\bibitem{goodfellow2017}
Ian Goodfellow, Nicolas Papernot, Sandy Huang, Yan Duan, Pieter Abbeel, and
  Jack Clark.
\newblock Attacking machine learning with adversarial examples, 2017.

\bibitem{goodfellow2014explaining}
Ian~J Goodfellow, Jonathon Shlens, and Christian Szegedy.
\newblock Explaining and harnessing adversarial examples.
\newblock {\em arXiv preprint arXiv:1412.6572}, 2014.

\bibitem{grosse17}
Kathrin Grosse, Nicolas Papernot, Praveen Manoharan, Michael Backes, and
  Patrick~D. McDaniel.
\newblock Adversarial examples for malware detection.
\newblock In {\em {ESORICS} 2017}, pages 62--79, 2017.

\bibitem{hassabis2017neuroscience}
Demis Hassabis, Dharshan Kumaran, Christopher Summerfield, and Matthew
  Botvinick.
\newblock Neuroscience-inspired artificial intelligence.
\newblock {\em Neuron}, 95(2):245--258, 2017.

\bibitem{resnet}
K.~{He}, X.~{Zhang}, S.~{Ren}, and J.~{Sun}.
\newblock {Identity Mappings in Deep Residual Networks}.
\newblock {\em ArXiv e-prints}, March 2016.

\bibitem{hillis2002combining}
James~M Hillis, Marc~O Ernst, Martin~S Banks, and Michael~S Landy.
\newblock Combining sensory information: mandatory fusion within, but not
  between, senses.
\newblock {\em Science}, 298(5598):1627--1630, 2002.

\bibitem{ibbotson2011visual}
Michael Ibbotson and Bart Krekelberg.
\newblock Visual perception and saccadic eye movements.
\newblock {\em Current opinion in neurobiology}, 21(4):553--558, 2011.

\bibitem{kolter2017provable}
J~Zico Kolter and Eric Wong.
\newblock Provable defenses against adversarial examples via the convex outer
  adversarial polytope.
\newblock {\em arXiv preprint arXiv:1711.00851}, 2017.

\bibitem{kovacs1995cortical}
GYULA KovAcs, Rufin Vogels, and Guy~A Orban.
\newblock Cortical correlate of pattern backward masking.
\newblock {\em Proceedings of the National Academy of Sciences},
  92(12):5587--5591, 1995.

\bibitem{kummerer2014deep}
Matthias K{\"u}mmerer, Lucas Theis, and Matthias Bethge.
\newblock Deep gaze i: Boosting saliency prediction with feature maps trained
  on imagenet.
\newblock {\em arXiv preprint arXiv:1411.1045}, 2014.

\bibitem{kummerer2017deepgaze}
Matthias K{\"u}mmerer, Tom Wallis, and Matthias Bethge.
\newblock Deepgaze ii: Predicting fixations from deep features over time and
  tasks.
\newblock {\em Journal of Vision}, 17(10):1147--1147, 2017.

\bibitem{kurakin2016mlatscale}
A.~{Kurakin}, I.~{Goodfellow}, and S.~{Bengio}.
\newblock {Adversarial Machine Learning at Scale}.
\newblock {\em ArXiv e-prints}, November 2016.

\bibitem{kurakin17physical}
Alexey Kurakin, Ian Goodfellow, and Samy Bengio.
\newblock Adversarial examples in the physical world.
\newblock In {\em ICLR'2017 Workshop}, 2016.

\bibitem{land2012animal}
Michael~F Land and Dan-Eric Nilsson.
\newblock {\em Animal eyes}.
\newblock Oxford University Press, 2012.

\bibitem{liu2016delving}
Yanpei Liu, Xinyun Chen, Chang Liu, and Dawn Song.
\newblock Delving into transferable adversarial examples and black-box attacks.
\newblock {\em arXiv preprint arXiv:1611.02770}, 2016.

\bibitem{madry2017towards}
Aleksander Madry, Aleksandar Makelov, Ludwig Schmidt, Dimitris Tsipras, and
  Adrian Vladu.
\newblock Towards deep learning models resistant to adversarial attacks.
\newblock {\em arXiv preprint arXiv:1706.06083}, 2017.

\bibitem{mcintosh2016deep}
Lane McIntosh, Niru Maheswaranathan, Aran Nayebi, Surya Ganguli, and Stephen
  Baccus.
\newblock Deep learning models of the retinal response to natural scenes.
\newblock In {\em Advances in neural information processing systems}, pages
  1369--1377, 2016.

\bibitem{olshausen201320}
Bruno~A Olshausen.
\newblock 20 years of learning about vision: Questions answered, questions
  unanswered, and questions not yet asked.
\newblock In {\em 20 Years of Computational Neuroscience}, pages 243--270.
  Springer, 2013.

\bibitem{papernot2016transferability}
Nicolas Papernot, Patrick McDaniel, and Ian Goodfellow.
\newblock Transferability in machine learning: from phenomena to black-box
  attacks using adversarial samples.
\newblock {\em arXiv preprint arXiv:1605.07277}, 2016.

\bibitem{papernot2017practical}
Nicolas Papernot, Patrick McDaniel, Ian Goodfellow, Somesh Jha, Z~Berkay Celik,
  and Ananthram Swami.
\newblock Practical black-box attacks against machine learning.
\newblock In {\em Proceedings of the 2017 ACM on Asia Conference on Computer
  and Communications Security}, pages 506--519. ACM, 2017.

\bibitem{papernot2016distillation}
Nicolas Papernot, Patrick McDaniel, Xi~Wu, Somesh Jha, and Ananthram Swami.
\newblock Distillation as a defense to adversarial perturbations against deep
  neural networks.
\newblock In {\em Security and Privacy (SP), 2016 IEEE Symposium on}, pages
  582--597. IEEE, 2016.

\bibitem{papernot2015limitations}
Nicolas Papernot, Patrick~D. McDaniel, Somesh Jha, Matt Fredrikson, Z.~Berkay
  Celik, and Ananthram Swami.
\newblock The limitations of deep learning in adversarial settings.
\newblock {\em CoRR}, abs/1511.07528, 2015.

\bibitem{potter2014detecting}
Mary~C Potter, Brad Wyble, Carl~Erick Hagmann, and Emily~S McCourt.
\newblock Detecting meaning in rsvp at 13 ms per picture.
\newblock {\em Attention, Perception, \& Psychophysics}, 76(2):270--279, 2014.

\bibitem{Rajalingham240614}
Rishi Rajalingham, Elias~B. Issa, Pouya Bashivan, Kohitij Kar, Kailyn Schmidt,
  and James~J DiCarlo.
\newblock Large-scale, high-resolution comparison of the core visual object
  recognition behavior of humans, monkeys, and state-of-the-art deep artificial
  neural networks.
\newblock {\em bioRxiv}, 2018.

\bibitem{riesenhuber1999hierarchical}
Maximilian Riesenhuber and Tomaso Poggio.
\newblock Hierarchical models of object recognition in cortex.
\newblock {\em Nature neuroscience}, 2(11):1019, 1999.

\bibitem{inceptionresnet}
C.~{Szegedy}, S.~{Ioffe}, V.~{Vanhoucke}, and A.~{Alemi}.
\newblock {Inception-v4, Inception-ResNet and the Impact of Residual
  Connections on Learning}.
\newblock {\em ArXiv e-prints}, February 2016.

\bibitem{inceptionv3}
C.~{Szegedy}, V.~{Vanhoucke}, S.~{Ioffe}, J.~{Shlens}, and Z.~{Wojna}.
\newblock {Rethinking the Inception Architecture for Computer Vision}.
\newblock {\em ArXiv e-prints}, December 2015.

\bibitem{szegedy2013intriguing}
Christian Szegedy, Wojciech Zaremba, Ilya Sutskever, Joan Bruna, Dumitru Erhan,
  Ian Goodfellow, and Rob Fergus.
\newblock Intriguing properties of neural networks.
\newblock {\em arXiv preprint arXiv:1312.6199}, 2013.

\bibitem{ensemble_training}
F.~{Tram{\`e}r}, A.~{Kurakin}, N.~{Papernot}, D.~{Boneh}, and P.~{McDaniel}.
\newblock {Ensemble Adversarial Training: Attacks and Defenses}.
\newblock {\em ArXiv e-prints}, May 2017.

\bibitem{retinal}
D.~C. Van~Essen and C.~H Anderson.
\newblock Information processing strategies and pathways in the primate visual
  system.
\newblock In Zornetzer~S. F., Davis~J. L., Lau C., and McKenna T., editors,
  {\em An introduction to neural and electronic networks}, page 45–76, San
  Diego, CA, 1995. Academic Press.

\bibitem{xu2017feature}
Weilin Xu, David Evans, and Yanjun Qi.
\newblock Feature squeezing: Detecting adversarial examples in deep neural
  networks.
\newblock {\em arXiv preprint arXiv:1704.01155}, 2017.

\bibitem{Yamins2016UsingGD}
Daniel L.~K. Yamins and James~J. DiCarlo.
\newblock Using goal-driven deep learning models to understand sensory cortex.
\newblock {\em Nature Neuroscience}, 19:356--365, 2016.

\end{thebibliography}
\bibliographystyle{plain}

\clearpage
\appendix

\normalsize

\part*{Supplemental material}

\setcounter{figure}{0} \renewcommand{\thefigure}{Supp.\arabic{figure}}
\setcounter{table}{0} \renewcommand{\thetable}{Supp.\arabic{table}}
\newcommand{\underscore}{$\_$}

\section{Supplementary Figures and Tables}
\label{sec: supp figures}

\begin{minipage}{\textwidth}
\begin{center}
\centerline{\includegraphics[width=\columnwidth]{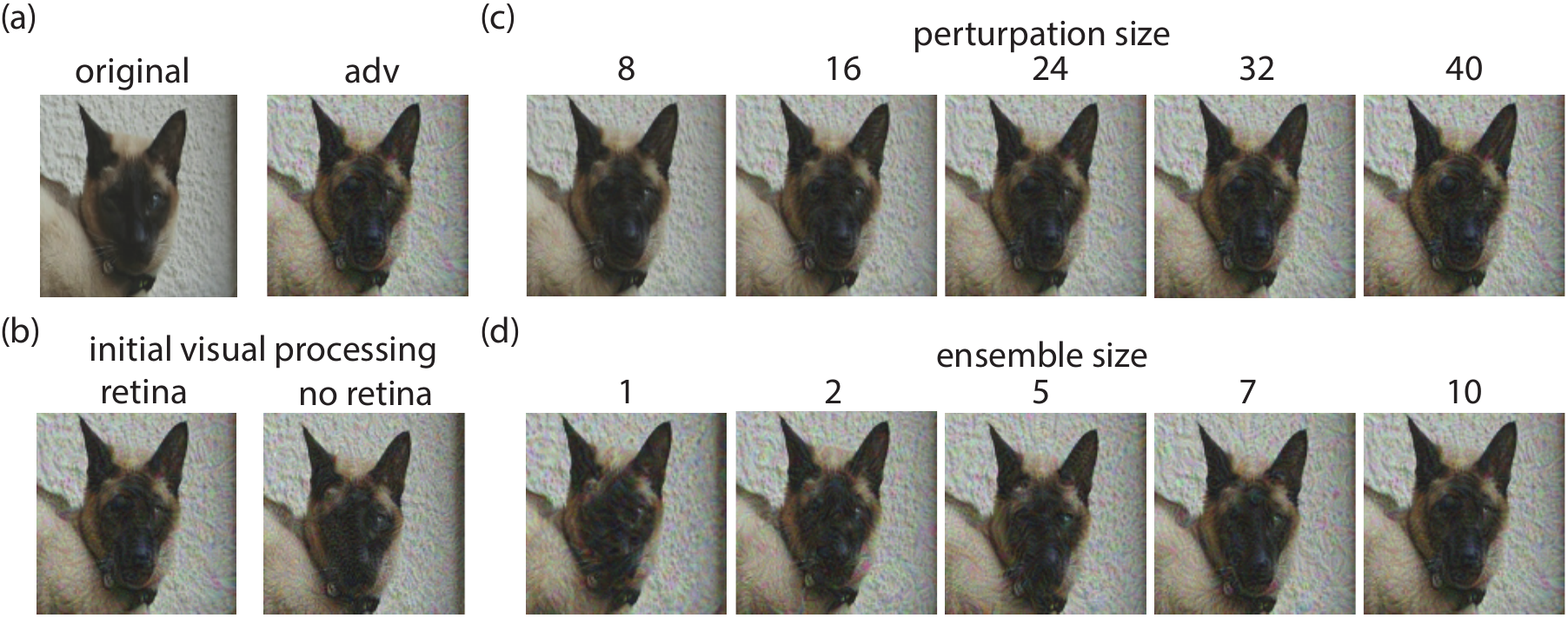}}
\captionof{figure}{\textbf{Intuition on factors contributing to the transfer to humans.} 
To give some intuition on the factors contributing to transfer, we examine a cat image from ImageNet ({\em (a)} left) that is already perceptually close to the target adversarial dog class, making the impact of subtle adversarial effects more obvious even on long observation ({\em (a)} right). 
Note that this image was not used in the experiment, and that typical images in the experiment {\em did not} fool unconstrained humans. 
{\em (b)} shows adversarial images with different perturbation sizes ranging from $\eps=8$ to $\eps=40$. 
Even smaller perturbation $\eps=8$ make the adversarial image perceptually more similar to a dog image, which suggests that transfer to humans may be robust to small $\epsilon$.
{\em (c)} Investigation of the importance of matching initial visual processing. The adversarial image on the left is similar to a dog, while removing the retina layer leads to an image which is less similar to a dog. This suggests that matching initial processing is an important factor in transferring adversarial examples to humans. 
{\em (d)}  Investigation of the importance of the number of models in the ensemble. We generated adversarial images with $\eps=32$ using an ensemble of size $1$-- $10$ models. One can see that adversarial perturbations become markedly less similar to a dog class as the number of models in the ensemble is reduced. 
This supports the importance of ensembling to the transfer of adversarial examples to humans.}
\label{fig cat dog}
\end{center}
\end{minipage}

\begin{minipage}{\textwidth}
\begin{center}
 \captionof{table}{\textbf{Adversarial examples transfer to humans.} Number of subjects that reported the correct class of images in the \texttt{adv} condition with lower mean accuracy compared to their mean accuracy in the \texttt{image} and \texttt{flip} conditions.}
 \begin{tabular}{C{0.6in} *2{C{.9in}} C{0.5in}}\toprule[1.5pt]
 \bf Group & \texttt{adv} $<$\texttt{image} & \texttt{adv} $<$ \texttt{flip} & \bf total \\\midrule

pets & 29   & 22  & 35 \\ 
hazard & 19  & 16 & 24  \\ 

 vegetables & 21 & 23 & 32  \\
\bottomrule[1.25pt]
\end{tabular}\\[1.5pt] \small{}
\label{table: people count}
\end{center}
\end{minipage}

\begin{minipage}{\textwidth}
\begin{center}
 \captionof{table}{\textbf{Accuracy of models on ImageNet validation set.} $^*$ models trained on ImageNet with retina layer pre-pended and with train data augmented with rescaled images in the range of [$40, 255-40$]; $^{**}$ model trained with adversarial examples augmented data. First ten models are models used in the adversarial training ensemble. Last two models are models used to test the transferability of adversarial examples.}

 \begin{tabular}{C{1.5in} *1{C{1.5in}}}\toprule[1.5pt]
 \bf Model  & \bf Top-1 accuracy \\\midrule

 Resnet V2 101 &         0.77  \\ 

 Resnet V2 101$^{*}$ &    0.7205  \\ 
 
 Inception V4 &         0.802  \\ 
 
 Inception V4$^{*}$ &         0.7518  \\

 Inception Resnet V2 &  0.804  \\ 

 Inception Resnet V2$^{*}$&  0.7662  \\ 

 Inception V3 &        0.78  \\ 
 
 Inception V3$^{*}$ &         0.7448  \\ 

 Resnet V2 152 &       0.778  \\

 Resnet V2 50$^{*}$ &      0.708  \\

 Resnet V2 50 (test) &  0.756  \\ 

 Inception V3$^{**}$ (test) & 0.776 \\ 
\bottomrule[1.25pt]
\end{tabular}\\[1.5pt] \small{}
\label{table: top 1 accuracy}
\end{center}
\end{minipage}

\begin{minipage}{\textwidth}
\begin{center}
 \captionof{table}{\textbf{Accuracy of ensemble used to generate adversarial examples on images at different conditions.} $^*$ models trained on ImageNet with retina layer appended and with train data augmented with rescaled images in the range of [$40, 255-40$]; 
 Numbers triplet reflects accuracy on images from pets, hazard, and vegetables groups, respectively.}
 \begin{tabular}{C{1.25in} *2{C{.7in}}}\toprule[1.5pt]
 \bf Train Model  & \texttt{adv} ($\%$)& \texttt{flip} ($\%$)\\\midrule

 Resnet V2 101 &         0.0, 0.0, 0.0 & 95, 92, 91  \\ 

 Resnet V2 101$^{*}$ &        0.0, 0.0, 0.0 & 87, 87, 77  \\ 
 
 Inception V4 &         0.0, 0.0, 0.0 & 96, 95, 86  \\ 
 
  Inception V4$^{*}$ &         0.0, 0.0, 0.0 & 87, 87, 73  \\

 Inception Resnet V2 & 0.0, 0.0, 0.0 & 97, 95, 95  \\ 
 
  Inception Resnet V2$^{*}$&  0.0, 0.0, 0.0 & 87, 83, 73  \\

 Inception V3 &         0.0, 0.0, 0.0 & 97, 94, 89  \\ 
 
 Inception V3$^{*}$ &         0.0, 0.0, 0.0 & 83, 86, 74  \\ 

 Resnet V2 152 &        0.0, 0.0, 0.0 & 96, 95, 91  \\

 Resnet V2 50$^{*}$ &        0.0, 0.0, 0.0 & 82, 85, 81  \\ 
\bottomrule[1.25pt]
\end{tabular}\\[1.5pt] 
\label{table: train accuracy}
\end{center}
\end{minipage}

\begin{minipage}{\textwidth}
\begin{center}
 \captionof{table}{\textbf{Accuracy of test models on images at different conditions.} $^{**}$ model trained on both clean and adversarial images. 
Numbers triplet is accuracy on  pets, hazard, and vegetables groups, respectively.}
 \begin{tabular}{C{1.25in} *2{C{.7in}}}\toprule[1.5pt]
 \bf Model  & \texttt{adv} ($\%$)& \texttt{flip} ($\%$)\\\midrule
 
  Resnet V2 50 & 8.7, 9.4, 13 & 93, 91, 85  \\ 

 Inception V3$^{**}$  & 6.0, 6.9, 17 & 95, 92, 94 \\
\bottomrule[1.25pt]
\end{tabular}\\[1.5pt] 
\label{table: test accuracy}
\end{center}
\end{minipage}

\begin{minipage}{\textwidth}
\begin{center}
 \captionof{table}{\textbf{Attack success on model ensemble.} Same convention as Table \ref{table: train accuracy}}
 \begin{tabular}{C{1.25in} *2{C{.78in}}}
 \toprule[1.5pt]
 \bf Model & \texttt{adv} ($\%$)& \texttt{flip} ($\%$)\\\midrule

 Resnet V2 101 &        100, 100, 100   & 2, 0, 0  \\ 

 Resnet V2 101$^{*}$ &       100, 100, 100  & 3, 0, 0  \\ 
 
 Inception V4 &         100, 100, 100  & 1, 0, 1  \\ 
 
 Inception V4$^{*}$ &       100, 100, 100  & 4, 1, 0  \\  
 
 Inception Resnet V2 &  100, 100, 100  & 1, 0, 1  \\ 

 Inception Resnet V2$^{*}$& 100, 100, 100  & 5, 2, 0  \\

 Inception V3 &         100, 100, 100  & 1, 0, 0  \\ 
 
 Inception V3$^{*}$ &       100, 100, 100  & 5, 1, 1  \\ 

 Resnet V2 152 &        100, 100, 100  & 1, 0, 0  \\

 Resnet V2 50$^{*}$ &       100, 100, 100  & 3, 1, 0  \\ 

\bottomrule[1.25pt]
\end{tabular}\\[1.5pt]
\label{table: train attack success}
\end{center}
\end{minipage}

\begin{minipage}{\textwidth}
\begin{center}
 \captionof{table}{\textbf{Attack success on test models.} 
  $^{**}$ model trained on both clean and adversarial images. Numbers triplet is error on pets, hazard, and vegetables groups, respectively.
 }
\centering
 \begin{tabular}{C{1.25in} *2{C{.78in}}}
 \toprule[1.5pt]
 \bf Model &   \texttt{adv} ($\%$)&   \texttt{flip} ($\%$)\\\midrule
 
  Resnet V2 50 & 87, 85, 57 & 1.3, 0.0, 0.0  \\ 

 Inception V3$^{**}$ & 89, 87, 74 & 1.5, 0.5, 0.0 \\ 
 
\bottomrule[1.25pt]
\end{tabular}\\[1.5pt]
\label{table: test attack success}
\vskip -0.1in
\end{center}
\end{minipage}

\section{Details of retinal blurring layer}\label{sec app retinal}

\subsection{Computing the primate eccentricity map}

Let $d_{viewer}$ be the distance (in meters) of the viewer from the display and $d_{hw}$ be the height and width of a square image (in meters). For every spatial position (in meters) $c = (x,y) \in R^2$ in the image 
we compute the retinal eccentricity (in radians) as follows:
\begin{align}
    \theta\left(c\right) &= \tan^{-1}(\frac{||c||_2}{d_{viewer}})
\end{align}
and turn this into a target resolution in units of radians
\begin{align}
    r_{rad}\left(c\right) &= \min\left( \alpha \theta\left(c\right), \beta \right)
.
\end{align}
We then turn this target resolution into a target spatial resolution in the plane of the screen,
\begin{align}
    r_m\left(c\right) &= r_{rad}\left(c\right) \left( 1 + \tan^2 \left(\theta(c)\right) \right)
, \\
    r_{pixel}\left(c\right) &= r_m\left(c\right) \cdot \text{[pixels per meter]}
.
\end{align}
This spatial resolution for two point discrimination is then converted into a corresponding low-pass cutoff frequency, in units of cycles per pixel,
\begin{align}
    f\left(c\right) &= \frac{\pi}{r_{pixel}}
,
\end{align}
where the numerator is $\pi$ rather than $2 \pi$ since the two point discrimination distance $r_{pixel}$ is half the wavelength.

Finally, this target low-pass spatial frequency $f\left(c\right)$ for each pixel is used to linearly interpolate each pixel value from the corresponding pixel in a set of low pass filtered images, as described in the following algorithm (all operations on matrices are assumed to be performed elementwise),
\begin{algorithm}
\caption{Applying retinal blur to an image}
\begin{algorithmic}[1]
    \State $X_{img} \gets \text{input image}$
    \State $F \gets \text{image containing corresponding target lowpass frequency for each pixel, computed from } f\left(c\right)$
    \State $\tilde{X} \gets \text{FFT}(X_{img})$
    \State $G \gets \text{norm of spatial frequency at each position in } Y$
    \State $\text{CUTOFF\_FREQS} \gets \text{list of frequencies to use as cutoffs for low-pass filtering}$
    \For{$f' \text{ in CUTOFF\_FREQS}$}
        \State $\tilde{Y}_{f'} \gets \tilde{X} \odot \exp\left( -\frac{G^2}{f^2} \right)$ 
        \State $Y_f \gets \text{InverseFFT}(\tilde{Y}_{f'})$
    \EndFor
    \State $w(c) \gets \text{linear interpolation coefficients for $F(c)$ into CUTOFF\_FREQS}  \quad \forall c$
    \State $X_{retinal}(c) \gets \sum_{f'} w_{f'}(c) Y_{f'}(c) \quad \forall c$
\end{algorithmic}
\end{algorithm}
We additionally cropped $X_{retinal}$ to $90\%$ width before use, to remove artifacts from the image edge.

Note that because the per-pixel blurring is performed using linear interpolation into images that were low-pass filtered in Fourier space, this transformation is both fast to compute and fully differentiable.

\section{Calculating probability of coarse class}
\label{sec: coarse prob}
To calculate the probability a model assigns to a coarse class, we summed probabilities assigned to the individual classes within the coarse class. Let $S_\mathrm{target}$ be the set of all individual labels in the target coarse class. Let $S_\mathrm{other}$ be all other individual labels not in the target coarse class. $|S_\mathrm{target}| +
|S_\mathrm{other}| = 1000$, since there are 1000 labels in ImageNet. Let $Y$ be the coarse class variable and $y_\mathrm{target}$ be our target coarse class. We can compute the probability a model $k$ assigns to a coarse class given image $X$ as
\begin{equation}
P_k(Y = y_\mathrm{target} | X) = \sum_{i \in S_\mathrm{target}} P_k(Y = y_i | X) = \sigma \left(\log \frac{\sum_{i \in S_\mathrm{target}}\tilde{F}_k\left(i | X\right)}{ \sum_{i \in S_\mathrm{other}} \tilde{F}_k\left(i | X \right)}\right)
\label{coarse_prob_eq}
\end{equation}
where $\tilde{F}_k\left(i | X \right)$ is the unnormalized probability assigned to fine class $i$ (in practice = $\text{exp}(logits)$ of class $i$). The coarse logit of the model with respect to the target class $y_\mathrm{target}$ is then $F_k (Y = y_\mathrm{target} | X) =\log \frac{\sum_{i \in S_\mathrm{target}}\tilde{F}_k\left(i | X\right)}{ \sum_{i \in S_\mathrm{other}} \tilde{F}_k\left(i | X \right)}$.

\section{Adversarial images generation.}
\label{sec: adv gen}
In the pipeline, an image is drawn from the source coarse class and perturbed to be classified as an image from the target coarse class. The attack method we use, the iterative targeted attack~\cite{kurakin17physical}, is performed as
\begin{align}
\tilde{X}_{adv}^n &= X_{adv}^{n-1} - \alpha * \mathrm{sign}( \nabla_{X^n}( J(X^n|y_\mathrm{target}) ) ),\nonumber \\
X_{adv}^{n} &= \mathrm{clip}\left( 
    \tilde{X}_{adv}^{n}, \left[ X - \epsilon, X + \epsilon\right]
\right),
\label{attack_eq}
\end{align}
where $J$ is the cost function as described below, $y_\mathrm{target}$ is the label of the target class, $\alpha$ is the step size, 
$X_{adv}^0 = X$ is the original clean image, and 
$X_{adv} = X^N_{adv}$ is the final adversarial image.
We set $\alpha = 2$, and $\epsilon$ is given per-condition in Section \ref{sec: experiment conditions}. 
After optimization, any perturbation whose $\ell_\infty$-norm was less than $\epsilon$ was scaled to have $\ell_\infty$-norm of $\epsilon$, for consistency across all perturbations.

Our goal was to create adversarial examples that transferred across many ML models before assessing their transferability to humans. To accomplish this, we created an ensemble from the geometric mean of several image classifiers, and performed the iterative attack on the ensemble loss~\citep{liu2016delving}
\begin{align}
J\left(X | y_{target}\right) &= - \log \left[P_\mathrm{ens} \left(y_\mathrm{target} | X\right) \right], \\
 P_\mathrm{ens} \left(y | X\right) &\propto
    \exp\left(\E_k{\left[\log  P_k \left(y | X\right)\right]}\right)
,
\label{cost_eq}
\end{align}
where $P_k\left( y | X\right)$ is the coarse class probabilities from model $k$, and $P_\mathrm{ens}\left( y | X \right)$ is the probability from the ensemble. In practice, $J\left(X | y_{target}\right)$ is equivalent to standard cross entropy loss based on coarse logits averaged across models in the ensemble (see Appendix \ref{sec: coarse prob} for the coarse logit definition).

To encourage a high transfer rate, we retained only adversarial examples that were successful against all 10 models for the {\tt adv} condition and at least $7/10$ models for the {\tt false} condition (see Section \ref{sec: experiment conditions} for condition definitions).

\section{Convolutional Neural Network Models}
\label{sec: models}

Some of the models in our ensemble are from a publicly available pretrained checkpoints\footnote{https://github.com/tensorflow/models/tree/master/research/slim}, and others are our own instances of the architectures, specifically trained for this experiment on ImageNet with the retinal layer prepended. To encourage invariance to image intensity scaling, we augmented each training batch with another batch with the same images but rescaled in the range of [$40, 255 - 40$], instead of [$0, 255$]. Supplementary Table \ref{table: top 1 accuracy} identifies all ten models used in the ensemble, and shows their top-1 accuracies, along with two holdout models that we used for evaluation.

\begin{minipage}{\textwidth}
\begin{center}
\fbox{%
    \parbox{\textwidth}{%
    \center
    ~ \\
        Image removed due to file size constraints. See \url{http://goo.gl/SJ8jpq} for full Supplemental Material with all images.\\ ~ \\
    }%
}
\captionof{figure}{\textbf{Adversarial Examples for false condition} (a) pets group. (b) hazard group. (c) vegetables group.}
\label{fig advx false}
\end{center}
\end{minipage}

\begin{minipage}{\textwidth}
\begin{center}
\fbox{%
    \parbox{\textwidth}{%
    \center
    ~ \\
        Image removed due to file size constraints. See \url{http://goo.gl/SJ8jpq} for full Supplemental Material with all images.\\ ~ \\
    }%
}
\captionof{figure}{\textbf{Adversarial Examples} pets group}
\label{fig advx pets}
\end{center}
\end{minipage}

\begin{minipage}{\textwidth}
\begin{center}
\fbox{%
    \parbox{\textwidth}{%
    \center
    ~ \\
        Image removed due to file size constraints. See \url{http://goo.gl/SJ8jpq} for full Supplemental Material with all images.\\ ~ \\
    }%
}
\captionof{figure}{\textbf{Adversarial Examples} hazard group}
\label{fig advx hazard}
\end{center}
\end{minipage}

\begin{minipage}{\textwidth}
\begin{center}
\fbox{%
    \parbox{\textwidth}{%
    \center
    ~ \\
        Image removed due to file size constraints. See \url{http://goo.gl/SJ8jpq} for full Supplemental Material with all images.\\ ~ \\
    }%
}
\captionof{figure}{\textbf{Adversarial Examples} vegetables group}
\label{fig advx veg}
\end{center}
\end{minipage}

\section{Image List from Imagenet}
\label{app image list}

The specific imagenet images used from each class in the experiments in this paper are as follows:

\textbf{dog:}

'n02106382\underscore564.JPEG', 'n02110958\underscore598.JPEG', 'n02101556\underscore13462.JPEG', 'n02113624\underscore7358.JPEG', 'n02113799\underscore2538.JPEG', 'n02091635\underscore11576.JPEG', 'n02106382\underscore2781.JPEG', 'n02112706\underscore105.JPEG', 'n02095570\underscore10951.JPEG', 'n02093859\underscore5274.JPEG', 'n02109525\underscore10825.JPEG', 'n02096294\underscore1400.JPEG', 'n02086646\underscore241.JPEG', 'n02098286\underscore5642.JPEG', 'n02106382\underscore9015.JPEG', 'n02090379\underscore9754.JPEG', 'n02102318\underscore10390.JPEG', 'n02086646\underscore4202.JPEG', 'n02086910\underscore5053.JPEG', 'n02113978\underscore3051.JPEG', 'n02093859\underscore3809.JPEG', 'n02105251\underscore2485.JPEG', 'n02109525\underscore35418.JPEG', 'n02108915\underscore7834.JPEG', 'n02113624\underscore430.JPEG', 'n02093256\underscore7467.JPEG', 'n02087046\underscore2701.JPEG', 'n02090379\underscore8849.JPEG', 'n02093754\underscore717.JPEG', 'n02086079\underscore15905.JPEG', 'n02102480\underscore4466.JPEG', 'n02107683\underscore5333.JPEG', 'n02102318\underscore8228.JPEG', 'n02099712\underscore867.JPEG', 'n02094258\underscore1958.JPEG', 'n02109047\underscore25075.JPEG', 'n02113624\underscore4304.JPEG', 'n02097474\underscore10985.JPEG', 'n02091032\underscore3832.JPEG', 
'n02085620\underscore859.JPEG', 'n02110806\underscore582.JPEG', 
'n02085782\underscore8327.JPEG', 'n02094258\underscore5318.JPEG', 'n02087046\underscore5721.JPEG', 'n02095570\underscore746.JPEG', 'n02099601\underscore3771.JPEG', 'n02102480\underscore41.JPEG', 'n02086910\underscore1048.JPEG', 'n02094114\underscore7299.JPEG', 'n02108551\underscore13160.JPEG', 'n02110185\underscore9847.JPEG', 'n02097298\underscore13025.JPEG', 'n02097298\underscore16751.JPEG', 'n02091467\underscore555.JPEG', 'n02113799\underscore2504.JPEG', 'n02085782\underscore14116.JPEG', 'n02097474\underscore13885.JPEG', 'n02105251\underscore8108.JPEG', 'n02113799\underscore3415.JPEG', 'n02095570\underscore8170.JPEG', 'n02088238\underscore1543.JPEG', 'n02097047\underscore6.JPEG', 'n02104029\underscore5268.JPEG', 'n02100583\underscore11473.JPEG', 'n02113978\underscore6888.JPEG', 'n02104365\underscore1737.JPEG', 'n02096177\underscore4779.JPEG', 'n02107683\underscore5303.JPEG', 'n02108915\underscore11155.JPEG', 'n02086910\underscore1872.JPEG', 'n02106550\underscore8383.JPEG', 'n02088094\underscore2191.JPEG', 'n02085620\underscore11897.JPEG', 'n02096051\underscore4802.JPEG', 'n02100735\underscore3641.JPEG', 'n02091032\underscore1389.JPEG', 'n02106382\underscore4671.JPEG', 'n02097298\underscore9059.JPEG', 'n02107312\underscore280.JPEG', 'n02111889\underscore86.JPEG', 'n02113978\underscore5397.JPEG', 'n02097209\underscore3461.JPEG', 'n02089867\underscore1115.JPEG', 'n02097658\underscore4987.JPEG', 'n02094114\underscore4125.JPEG', 'n02100583\underscore130.JPEG', 'n02112137\underscore5859.JPEG', 'n02113799\underscore19636.JPEG', 'n02088094\underscore5488.JPEG', 'n02089078\underscore393.JPEG', 'n02098413\underscore1794.JPEG', 'n02113799\underscore1970.JPEG', 'n02091032\underscore3655.JPEG', 'n02105855\underscore11127.JPEG', 'n02096294\underscore3025.JPEG', 'n02094114\underscore4831.JPEG', 'n02111889\underscore10472.JPEG', 'n02113624\underscore9125.JPEG', 'n02097474\underscore9719.JPEG', 'n02094433\underscore2451.JPEG', 'n02095889\underscore6464.JPEG', 'n02093256\underscore458.JPEG', 'n02091134\underscore2732.JPEG', 'n02091244\underscore2622.JPEG', 'n02094114\underscore2169.JPEG', 'n02090622\underscore2337.JPEG', 'n02101556\underscore6764.JPEG', 'n02096051\underscore1459.JPEG', 'n02087046\underscore9056.JPEG', 'n02098105\underscore8405.JPEG', 'n02112137\underscore5696.JPEG', 'n02110806\underscore7949.JPEG', 'n02097298\underscore2420.JPEG', 'n02085620\underscore6814.JPEG', 'n02108915\underscore1703.JPEG', 'n02100877\underscore19273.JPEG', 'n02106550\underscore3765.JPEG', 'n02107312\underscore3524.JPEG', 'n02111889\underscore2963.JPEG', 'n02113624\underscore9129.JPEG', 'n02097047\underscore3200.JPEG', 
'n02093256\underscore8365.JPEG', 'n02093991\underscore9420.JPEG', 'n02112137\underscore1635.JPEG', 'n02111129\underscore3530.JPEG', 'n02101006\underscore8123.JPEG', 'n02102040\underscore5033.JPEG', 'n02113624\underscore437.JPEG', 'n02090622\underscore5866.JPEG', 'n02110806\underscore3711.JPEG', 'n02112137\underscore14788.JPEG', 'n02105162\underscore7406.JPEG', 'n02097047\underscore5061.JPEG', 'n02108422\underscore11587.JPEG', 'n02091467\underscore4265.JPEG', 'n02091467\underscore12683.JPEG', 'n02104365\underscore3628.JPEG', 'n02086646\underscore3314.JPEG', 'n02099849\underscore736.JPEG', 'n02100735\underscore8112.JPEG', 'n02112018\underscore12764.JPEG', 'n02093428\underscore11175.JPEG', 'n02110627\underscore9822.JPEG', 'n02107142\underscore24318.JPEG', 'n02105162\underscore5489.JPEG', 'n02093754\underscore5904.JPEG', 'n02110958\underscore215.JPEG', 'n02095314\underscore4027.JPEG', 'n02109961\underscore3250.JPEG', 'n02108551\underscore7343.JPEG', 'n02110627\underscore10272.JPEG', 'n02088364\underscore3099.JPEG', 'n02110806\underscore2721.JPEG', 'n02095314\underscore2261.JPEG', 'n02106550\underscore9870.JPEG', 'n02107574\underscore3991.JPEG', 'n02095570\underscore3288.JPEG', 'n02086079\underscore39042.JPEG', 'n02096294\underscore9416.JPEG', 'n02110806\underscore6528.JPEG', 'n02088466\underscore11397.JPEG', 'n02092002\underscore996.JPEG', 'n02098413\underscore8605.JPEG', 'n02085620\underscore712.JPEG', 'n02100236\underscore3011.JPEG', 'n02086646\underscore7788.JPEG', 'n02085620\underscore4661.JPEG', 'n02098105\underscore1746.JPEG', 'n02113624\underscore8608.JPEG', 'n02097474\underscore1168.JPEG', 'n02107683\underscore1496.JPEG', 'n02110185\underscore12849.JPEG', 'n02085620\underscore11946.JPEG', 'n02087394\underscore16385.JPEG', 'n02110806\underscore22671.JPEG', 'n02113624\underscore526.JPEG', 'n02096294\underscore12642.JPEG', 'n02113023\underscore7510.JPEG', 'n02088364\underscore13285.JPEG', 'n02095889\underscore2977.JPEG', 'n02105056\underscore9215.JPEG', 'n02102318\underscore9744.JPEG', 'n02097298\underscore11834.JPEG', 'n02111277\underscore16201.JPEG', 'n02085782\underscore8518.JPEG', 'n02113978\underscore11280.JPEG', 'n02106382\underscore10700.JPEG'.

\textbf{cat:}

'n02123394\underscore661.JPEG', 'n02123045\underscore11954.JPEG', 'n02123394\underscore3695.JPEG', 'n02123394\underscore2692.JPEG', 'n02123597\underscore12166.JPEG', 'n02123045\underscore7014.JPEG', 'n02123159\underscore2777.JPEG', 'n02123394\underscore684.JPEG', 'n02124075\underscore543.JPEG', 'n02123597\underscore7557.JPEG', 'n02124075\underscore7857.JPEG', 'n02123597\underscore3770.JPEG', 'n02124075\underscore4986.JPEG', 'n02123045\underscore568.JPEG', 'n02123394\underscore1541.JPEG', 'n02123597\underscore3498.JPEG', 'n02123597\underscore10304.JPEG', 'n02123394\underscore2084.JPEG', 'n02123597\underscore5283.JPEG', 'n02123597\underscore13807.JPEG', 'n02124075\underscore12282.JPEG', 'n02123597\underscore8575.JPEG', 'n02123045\underscore11787.JPEG', 'n02123394\underscore888.JPEG', 'n02123045\underscore1815.JPEG', 'n02123394\underscore7614.JPEG', 'n02123597\underscore27865.JPEG', 'n02124075\underscore1279.JPEG', 'n02123394\underscore4775.JPEG', 'n02123394\underscore976.JPEG', 'n02123394\underscore8385.JPEG', 'n02123597\underscore14791.JPEG', 'n02123045\underscore10424.JPEG', 'n02123597\underscore7698.JPEG', 'n02124075\underscore8140.JPEG', 'n02123045\underscore3754.JPEG', 'n02123597\underscore1819.JPEG', 'n02123597\underscore395.JPEG', 'n02123394\underscore415.JPEG', 'n02124075\underscore9747.JPEG', 'n02123045\underscore9467.JPEG', 'n02123159\underscore6842.JPEG', 'n02123394\underscore9611.JPEG', 'n02123597\underscore7283.JPEG', 'n02123597\underscore11799.JPEG', 'n02123597\underscore660.JPEG', 'n02123045\underscore7511.JPEG', 'n02123597\underscore10723.JPEG', 'n02123159\underscore7836.JPEG', 'n02123597\underscore14530.JPEG', 'n02123597\underscore28555.JPEG', 'n02123394\underscore6079.JPEG', 'n02123394\underscore6792.JPEG', 'n02123597\underscore11564.JPEG', 'n02123597\underscore8916.JPEG', 
'n02124075\underscore123.JPEG', 'n02123045\underscore5150.JPEG', 'n02124075\underscore353.JPEG', 'n02123597\underscore12941.JPEG', 'n02123045\underscore10095.JPEG', 'n02123597\underscore6533.JPEG', 'n02123045\underscore4611.JPEG', 'n02123597\underscore754.JPEG', 'n02123394\underscore8561.JPEG', 'n02123597\underscore6409.JPEG', 'n02123159\underscore4909.JPEG', 'n02123597\underscore564.JPEG', 'n02123394\underscore1633.JPEG', 'n02123394\underscore1196.JPEG', 'n02123394\underscore2787.JPEG', 'n02124075\underscore10542.JPEG', 'n02123597\underscore6242.JPEG', 'n02123597\underscore3063.JPEG', 'n02123597\underscore13164.JPEG', 'n02123045\underscore7449.JPEG', 'n02123045\underscore13299.JPEG', 'n02123394\underscore8165.JPEG', 'n02123394\underscore1852.JPEG', 'n02123597\underscore8771.JPEG', 'n02123159\underscore6581.JPEG', 'n02123394\underscore5906.JPEG', 'n02124075\underscore2747.JPEG', 'n02124075\underscore11383.JPEG', 'n02123597\underscore3919.JPEG', 'n02123394\underscore2514.JPEG', 'n02124075\underscore7423.JPEG', 'n02123394\underscore6968.JPEG', 'n02123045\underscore4850.JPEG', 'n02123045\underscore10689.JPEG', 'n02124075\underscore13539.JPEG', 'n02123597\underscore13378.JPEG', 'n02123159\underscore4847.JPEG', 'n02123394\underscore1798.JPEG', 'n02123597\underscore27951.JPEG', 'n02123159\underscore587.JPEG', 'n02123597\underscore1825.JPEG', 'n02123159\underscore2200.JPEG', 'n02123597\underscore12.JPEG', 'n02123597\underscore6778.JPEG', 'n02123597\underscore6693.JPEG', 'n02123045\underscore11782.JPEG', 'n02123597\underscore13706.JPEG', 'n02123394\underscore9032.JPEG', 'n02124075\underscore4459.JPEG', 'n02123597\underscore13752.JPEG', 'n02123394\underscore2285.JPEG', 'n02123597\underscore1410.JPEG', 'n02123159\underscore6134.JPEG', 'n02123597\underscore11290.JPEG', 'n02123597\underscore6347.JPEG', 'n02123394\underscore1789.JPEG', 
'n02123045\underscore11255.JPEG', 'n02123394\underscore6096.JPEG', 'n02123394\underscore4081.JPEG', 'n02123394\underscore5679.JPEG', 'n02123394\underscore2471.JPEG', 'n02123159\underscore5797.JPEG', 'n02123597\underscore13894.JPEG', 'n02124075\underscore10854.JPEG', 'n02123394\underscore8605.JPEG', 'n02124075\underscore8281.JPEG', 'n02123597\underscore11724.JPEG', 'n02123394\underscore8242.JPEG', 'n02123394\underscore3569.JPEG', 'n02123597\underscore10639.JPEG', 'n02123045\underscore3818.JPEG', 'n02124075\underscore6459.JPEG', 'n02123394\underscore185.JPEG', 'n02123597\underscore8961.JPEG', 'n02124075\underscore9743.JPEG', 'n02123394\underscore1627.JPEG', 'n02123597\underscore13175.JPEG', 'n02123045\underscore2694.JPEG', 'n02123597\underscore4537.JPEG', 'n02123597\underscore6400.JPEG', 'n02123045\underscore7423.JPEG', 'n02123597\underscore3004.JPEG', 'n02123394\underscore2988.JPEG', 'n02124075\underscore9512.JPEG', 'n02123394\underscore6318.JPEG', 'n02123597\underscore1843.JPEG', 'n02124075\underscore2053.JPEG', 'n02123597\underscore3828.JPEG', 'n02123394\underscore14.JPEG', 'n02123394\underscore8141.JPEG', 'n02124075\underscore1624.JPEG', 'n02123597\underscore459.JPEG', 'n02124075\underscore6405.JPEG', 'n02123045\underscore8595.JPEG', 'n02123159\underscore3226.JPEG', 'n02124075\underscore9141.JPEG', 'n02123597\underscore2031.JPEG', 'n02123045\underscore2354.JPEG', 'n02123597\underscore6710.JPEG', 'n02123597\underscore6613.JPEG', 'n02123159\underscore1895.JPEG', 'n02123394\underscore2953.JPEG', 'n02123394\underscore5846.JPEG', 'n02123394\underscore513.JPEG', 'n02123045\underscore16637.JPEG', 'n02123394\underscore7848.JPEG', 'n02123394\underscore3229.JPEG', 'n02123045\underscore8881.JPEG', 'n02123394\underscore8250.JPEG', 'n02124075\underscore7651.JPEG', 'n02123394\underscore200.JPEG', 'n02123394\underscore2814.JPEG', 'n02123045\underscore6445.JPEG', 'n02123394\underscore2467.JPEG', 'n02123045\underscore3317.JPEG', 'n02123597\underscore1422.JPEG', 'n02123597\underscore13442.JPEG', 'n02123394\underscore8225.JPEG', 'n02123597\underscore9337.JPEG', 'n02123394\underscore32.JPEG', 'n02123394\underscore2193.JPEG', 'n02123394\underscore1625.JPEG', 'n02123597\underscore8799.JPEG', 'n02123597\underscore13241.JPEG', 'n02123597\underscore7681.JPEG', 'n02123597\underscore4550.JPEG', 'n02123597\underscore3896.JPEG', 'n02123394\underscore9554.JPEG', 'n02124075\underscore13600.JPEG', 'n02123394\underscore571.JPEG', 'n02123597\underscore10886.JPEG', 'n02123045\underscore6741.JPEG',
'n02123045\underscore10438.JPEG', 'n02123045\underscore9954.JPEG'.

\textbf{spider:}

'n01775062\underscore517.JPEG', 'n01774750\underscore18017.JPEG', 'n01774384\underscore13186.JPEG', 'n01774750\underscore3115.JPEG', 'n01775062\underscore5075.JPEG', 'n01773549\underscore1541.JPEG', 'n01775062\underscore4867.JPEG', 'n01775062\underscore8156.JPEG', 'n01774750\underscore7128.JPEG', 'n01775062\underscore4632.JPEG', 'n01773549\underscore8734.JPEG', 'n01773549\underscore2274.JPEG', 'n01773549\underscore10298.JPEG', 'n01774384\underscore1811.JPEG', 'n01774750\underscore7498.JPEG', 'n01774750\underscore10265.JPEG', 'n01773549\underscore1964.JPEG', 'n01774750\underscore3268.JPEG', 'n01773549\underscore6095.JPEG', 'n01775062\underscore8812.JPEG', 'n01774750\underscore10919.JPEG', 'n01775062\underscore1180.JPEG', 'n01773549\underscore7275.JPEG', 'n01773549\underscore9346.JPEG', 'n01773549\underscore8243.JPEG', 'n01775062\underscore3127.JPEG', 'n01773549\underscore10608.JPEG', 'n01773549\underscore3442.JPEG', 'n01773157\underscore1487.JPEG', 'n01774750\underscore7775.JPEG', 'n01775062\underscore419.JPEG', 'n01774750\underscore7638.JPEG', 'n01775062\underscore847.JPEG', 'n01774750\underscore3154.JPEG', 'n01773549\underscore1534.JPEG', 'n01773157\underscore1039.JPEG', 'n01775062\underscore5644.JPEG', 'n01775062\underscore8525.JPEG', 'n01773797\underscore216.JPEG', 'n01775062\underscore900.JPEG', 'n01774750\underscore8513.JPEG', 'n01774750\underscore3424.JPEG', 'n01774750\underscore3085.JPEG', 'n01775062\underscore3662.JPEG', 'n01774384\underscore15681.JPEG', 'n01774750\underscore326.JPEG', 'n01773157\underscore9503.JPEG', 'n01774750\underscore3332.JPEG', 'n01774750\underscore2799.JPEG', 'n01773157\underscore10606.JPEG', 'n01773157\underscore1905.JPEG', 'n01773549\underscore379.JPEG', 'n01773797\underscore597.JPEG', 'n01773157\underscore3226.JPEG', 'n01774750\underscore7875.JPEG', 'n01774384\underscore16102.JPEG', 'n01773549\underscore2832.JPEG', 'n01775062\underscore5072.JPEG', 'n01773549\underscore4278.JPEG', 'n01773549\underscore5854.JPEG', 'n01774384\underscore1998.JPEG', 'n01774750\underscore13875.JPEG', 'n01775062\underscore8270.JPEG', 'n01773549\underscore2941.JPEG', 'n01774750\underscore5235.JPEG', 'n01773549\underscore4150.JPEG', 'n01774750\underscore6217.JPEG', 'n01775062\underscore3137.JPEG', 'n01774750\underscore5480.JPEG', 'n01774384\underscore11955.JPEG', 'n01775062\underscore8376.JPEG', 'n01773157\underscore2688.JPEG', 'n01773549\underscore6825.JPEG', 'n01774750\underscore10422.JPEG', 'n01774384\underscore20786.JPEG', 'n01773549\underscore398.JPEG', 'n01773549\underscore4965.JPEG', 'n01774750\underscore7470.JPEG', 'n01775062\underscore1379.JPEG', 'n01774384\underscore2399.JPEG', 'n01773549\underscore9799.JPEG', 'n01775062\underscore305.JPEG', 'n01774384\underscore15519.JPEG', 'n01774750\underscore3333.JPEG', 'n01774750\underscore2604.JPEG', 'n01774750\underscore3134.JPEG', 'n01774750\underscore4646.JPEG', 'n01775062\underscore5009.JPEG', 'n01774750\underscore10200.JPEG', 'n01775062\underscore7964.JPEG', 'n01774384\underscore2458.JPEG', 'n01773797\underscore3333.JPEG', 'n01774750\underscore9987.JPEG', 'n01773549\underscore5790.JPEG', 'n01773549\underscore854.JPEG', 'n01774750\underscore11370.JPEG', 'n01774750\underscore10698.JPEG', 'n01774750\underscore9287.JPEG', 'n01773797\underscore6703.JPEG', 'n01773797\underscore931.JPEG', 'n01773549\underscore5280.JPEG', 'n01773797\underscore5385.JPEG', 'n01773797\underscore1098.JPEG', 'n01774750\underscore436.JPEG', 'n01774384\underscore13770.JPEG', 'n01774750\underscore9780.JPEG', 'n01774750\underscore8640.JPEG', 'n01774750\underscore653.JPEG', 'n01774384\underscore12554.JPEG', 'n01774750\underscore9716.JPEG'

\textbf{snake:}

'n01737021\underscore7081.JPEG', 'n01728572\underscore16119.JPEG', 'n01735189\underscore10620.JPEG', 'n01751748\underscore3573.JPEG', 'n01729322\underscore6690.JPEG', 'n01735189\underscore20703.JPEG', 'n01734418\underscore4792.JPEG', 'n01749939\underscore2784.JPEG', 'n01729977\underscore4113.JPEG', 'n01756291\underscore6505.JPEG', 'n01742172\underscore3003.JPEG', 'n01728572\underscore19317.JPEG', 'n01739381\underscore5838.JPEG', 'n01737021\underscore1381.JPEG', 'n01749939\underscore4704.JPEG', 'n01755581\underscore10792.JPEG', 'n01729977\underscore9474.JPEG', 'n01744401\underscore11909.JPEG', 'n01739381\underscore10303.JPEG', 'n01749939\underscore820.JPEG', 'n01728572\underscore27743.JPEG', 'n01734418\underscore12057.JPEG', 'n01742172\underscore8636.JPEG', 'n01729977\underscore14112.JPEG', 'n01739381\underscore6286.JPEG', 'n01734418\underscore761.JPEG', 'n01740131\underscore13437.JPEG', 'n01728920\underscore9571.JPEG', 'n01753488\underscore4234.JPEG', 'n01749939\underscore5712.JPEG', 'n01739381\underscore6072.JPEG', 'n01739381\underscore7683.JPEG', 'n01729322\underscore9202.JPEG', 'n01751748\underscore13413.JPEG', 'n01756291\underscore4626.JPEG', 'n01742172\underscore9733.JPEG', 'n01737021\underscore12610.JPEG', 'n01739381\underscore87.JPEG', 'n01729977\underscore1134.JPEG', 'n01753488\underscore637.JPEG', 'n01748264\underscore18478.JPEG', 'n01728572\underscore22360.JPEG', 'n01737021\underscore3386.JPEG', 'n01751748\underscore560.JPEG', 'n01751748\underscore18223.JPEG', 'n01749939\underscore5750.JPEG', 'n01748264\underscore7044.JPEG', 'n01739381\underscore1163.JPEG', 'n01751748\underscore311.JPEG', 'n01756291\underscore9028.JPEG', 'n01739381\underscore10473.JPEG', 'n01728572\underscore1415.JPEG', 'n01729322\underscore10918.JPEG', 'n01748264\underscore653.JPEG', 'n01753488\underscore10957.JPEG', 'n01756291\underscore3990.JPEG', 'n01756291\underscore11915.JPEG', 'n01756291\underscore6776.JPEG', 'n01740131\underscore11661.JPEG', 'n01729977\underscore5715.JPEG', 'n01737021\underscore16733.JPEG', 'n01753488\underscore15197.JPEG', 'n01744401\underscore7248.JPEG', 'n01728572\underscore7661.JPEG', 'n01740131\underscore13680.JPEG', 'n01729322\underscore5446.JPEG', 'n01749939\underscore6508.JPEG', 'n01748264\underscore2140.JPEG', 'n01729977\underscore16782.JPEG', 'n01748264\underscore7602.JPEG', 'n01756291\underscore17857.JPEG', 'n01729977\underscore461.JPEG', 'n01742172\underscore20552.JPEG', 'n01735189\underscore3258.JPEG', 'n01728920\underscore9265.JPEG', 'n01748264\underscore18133.JPEG', 'n01748264\underscore16699.JPEG', 'n01739381\underscore1006.JPEG', 'n01753488\underscore10555.JPEG', 'n01751748\underscore3202.JPEG', 'n01734418\underscore3929.JPEG', 'n01751748\underscore5908.JPEG', 'n01751748\underscore8470.JPEG', 'n01739381\underscore3598.JPEG', 'n01739381\underscore255.JPEG', 'n01729977\underscore15657.JPEG', 'n01748264\underscore21477.JPEG', 'n01751748\underscore2912.JPEG', 'n01728920\underscore9154.JPEG', 'n01728572\underscore17552.JPEG', 'n01740131\underscore14560.JPEG', 'n01729322\underscore5947.JPEG'.

\textbf{Broccoli:}

'n07714990\underscore8640.JPEG', 'n07714990\underscore5643.JPEG', 'n07714990\underscore7777.JPEG', 'n07714990\underscore888.JPEG', 'n07714990\underscore3398.JPEG', 'n07714990\underscore4576.JPEG', 'n07714990\underscore8554.JPEG', 'n07714990\underscore1957.JPEG', 'n07714990\underscore4201.JPEG', 'n07714990\underscore3130.JPEG', 'n07714990\underscore4115.JPEG', 'n07714990\underscore524.JPEG', 'n07714990\underscore6504.JPEG', 'n07714990\underscore3125.JPEG', 'n07714990\underscore5838.JPEG', 'n07714990\underscore1779.JPEG', 'n07714990\underscore6393.JPEG', 'n07714990\underscore1409.JPEG', 'n07714990\underscore4962.JPEG', 'n07714990\underscore7282.JPEG', 'n07714990\underscore7314.JPEG', 'n07714990\underscore11933.JPEG', 'n07714990\underscore1202.JPEG', 'n07714990\underscore3626.JPEG', 'n07714990\underscore7873.JPEG', 'n07714990\underscore3325.JPEG', 'n07714990\underscore3635.JPEG', 'n07714990\underscore12524.JPEG', 'n07714990\underscore14952.JPEG', 'n07714990\underscore7048.JPEG', 'n07714990\underscore500.JPEG', 'n07714990\underscore7950.JPEG', 'n07714990\underscore2445.JPEG', 'n07714990\underscore1294.JPEG', 'n07714990\underscore7336.JPEG', 'n07714990\underscore14743.JPEG', 'n07714990\underscore1423.JPEG', 'n07714990\underscore2185.JPEG', 'n07714990\underscore6566.JPEG', 'n07714990\underscore567.JPEG', 'n07714990\underscore1532.JPEG', 'n07714990\underscore5212.JPEG', 'n07714990\underscore8971.JPEG', 'n07714990\underscore6116.JPEG', 'n07714990\underscore5462.JPEG', 'n07714990\underscore7644.JPEG', 'n07714990\underscore8596.JPEG', 'n07714990\underscore1138.JPEG', 'n07714990\underscore15078.JPEG', 'n07714990\underscore1602.JPEG', 'n07714990\underscore2460.JPEG', 'n07714990\underscore159.JPEG', 'n07714990\underscore9445.JPEG', 'n07714990\underscore471.JPEG', 'n07714990\underscore1777.JPEG', 'n07714990\underscore9760.JPEG', 'n07714990\underscore1528.JPEG', 'n07714990\underscore12338.JPEG', 'n07714990\underscore2201.JPEG', 'n07714990\underscore6850.JPEG', 'n07714990\underscore4492.JPEG', 'n07714990\underscore7791.JPEG', 'n07714990\underscore9752.JPEG', 'n07714990\underscore1702.JPEG', 'n07714990\underscore3682.JPEG', 'n07714990\underscore14342.JPEG', 'n07714990\underscore2661.JPEG', 'n07714990\underscore5467.JPEG'. $~~~~~~~~~~~~~~~~~~~~~~~~~~~~~~~~~~~~~~~~~$

\textbf{Cabbage:}

'n07714571\underscore14784.JPEG', 'n07714571\underscore4795.JPEG', 'n07714571\underscore11969.JPEG', 'n07714571\underscore1394.JPEG', 'n07714571\underscore4155.JPEG', 'n07714571\underscore3624.JPEG', 'n07714571\underscore13753.JPEG', 'n07714571\underscore7351.JPEG', 'n07714571\underscore10316.JPEG', 'n07714571\underscore7235.JPEG', 'n07714571\underscore17716.JPEG', 'n07714571\underscore1639.JPEG', 'n07714571\underscore5107.JPEG', 'n07714571\underscore4109.JPEG', 'n07714571\underscore11878.JPEG', 'n07714571\underscore15910.JPEG', 'n07714571\underscore14401.JPEG', 'n07714571\underscore2741.JPEG', 'n07714571\underscore8576.JPEG', 'n07714571\underscore1624.JPEG', 'n07714571\underscore13479.JPEG', 'n07714571\underscore2715.JPEG', 'n07714571\underscore3676.JPEG', 'n07714571\underscore12371.JPEG', 'n07714571\underscore4829.JPEG', 'n07714571\underscore3922.JPEG', 'n07714571\underscore10377.JPEG', 'n07714571\underscore8040.JPEG', 'n07714571\underscore8147.JPEG', 'n07714571\underscore10377.JPEG', 'n07714571\underscore8040.JPEG', 'n07714571\underscore5730.JPEG', 'n07714571\underscore16460.JPEG', 'n07714571\underscore8198.JPEG', 'n07714571\underscore1095.JPEG', 'n07714571\underscore3922.JPEG', 'n07714571\underscore7745.JPEG', 'n07714571\underscore6301.JPEG'.

\end{document}